\documentclass[journal,comsoc]{IEEEtran}
\usepackage[T1]{fontenc}

\ifCLASSINFOpdf
   \usepackage{graphicx}
  \graphicspath{{imgs/}}
\else
\fi
\usepackage{color}
\usepackage{amsmath}
\usepackage[cmintegrals]{newtxmath}
\begin{document}
%
\title{Dense Registration and Mosaicking of Fingerprints\\ by Training an End-to-End Network}
%
%
%

\author{Zhe~Cui,
        Jianjiang~Feng,~\IEEEmembership{Member,~IEEE,}
        Jie~Zhou,~\IEEEmembership{Senior~Member,~IEEE}
\thanks{

The authors are with the Department of Automation, the State Key Lab of Intelligent Technologies and Systems, and Beijing National Research Center for Information Science and Technology (BNRist), Tsinghua University, Beijing 100084, China.

e-mail: cuiz17@mails.tsinghua.edu.cn, jfeng@tsinghua.edu.cn, jzhou@tsinghua.edu.cn.

}
}

%
%

\markboth{IEEE TRANSACTIONS ON INFORMATION FORENSICS AND SECURITY}%
{Shell \MakeLowercase{\textit{et al.}}: Bare Demo of IEEEtran.cls for IEEE Communications Society Journals}
%



\maketitle

\begin{abstract}
Dense registration of fingerprints is a challenging task due to elastic skin distortion, low image quality, and self-similarity of ridge pattern. To overcome the limitation of handcraft features, we propose to train an end-to-end network to directly output pixel-wise displacement field between two fingerprints. The proposed network includes a siamese network for feature embedding, and a following encoder-decoder network for regressing displacement field. By applying displacement fields reliably estimated by tracing high quality fingerprint videos to challenging fingerprints, we synthesize a large number of training fingerprint pairs with ground truth displacement fields. In addition, based on the proposed registration algorithm, we propose a fingerprint mosaicking method based on optimal seam selection. Registration and matching experiments on FVC2004 databases, Tsinghua Distorted Fingerprint (TDF) database, and NIST SD27 latent fingerprint database show that our registration method outperforms previous dense registration methods in accuracy and efficiency. Mosaicking experiment on FVC2004 DB1 demonstrates that the proposed algorithm produced higher quality fingerprints than other algorithms which also validates the performance of our registration algorithm.
\end{abstract}


\section{Introduction}\label{sec:introduction}
Although automatic fingerprint recognition system has been widely deployed in various applications, current fingerprint matching algorithms still need improvement, especially for those fingerprints with large skin distortion \cite{2009handbook}. As an inherent problem in contact-based fingerprint acquisition, skin distortion increases intra-class variations among different images of a same finger, and thus causes declines in fingerprint matching accuracy \cite{Cappelli2006}. 

Fingerprint registration algorithms can be employed to reduce negative impact of skin distortion. Conventional fingerprint registration algorithms \cite{Bazen2003}\cite{Tico2003}\cite{Ross2005} typically first find minutiae correspondences between two fingerprints (referred to as input fingerprint and reference fingerprint), and then fit a spatial transformation model to these corresponding minutiae. However, for highly distorted fingerprints, minutiae-based registration methods can only obtain a very sparse set of displacement measures at locations of corresponding minutiae, which cannot align all ridges in two fingerprints. 

\begin{figure*}[htb]
\centering
\begin{minipage}[c]{.3\linewidth}
  \centering
  \centerline{\includegraphics[width=\linewidth]{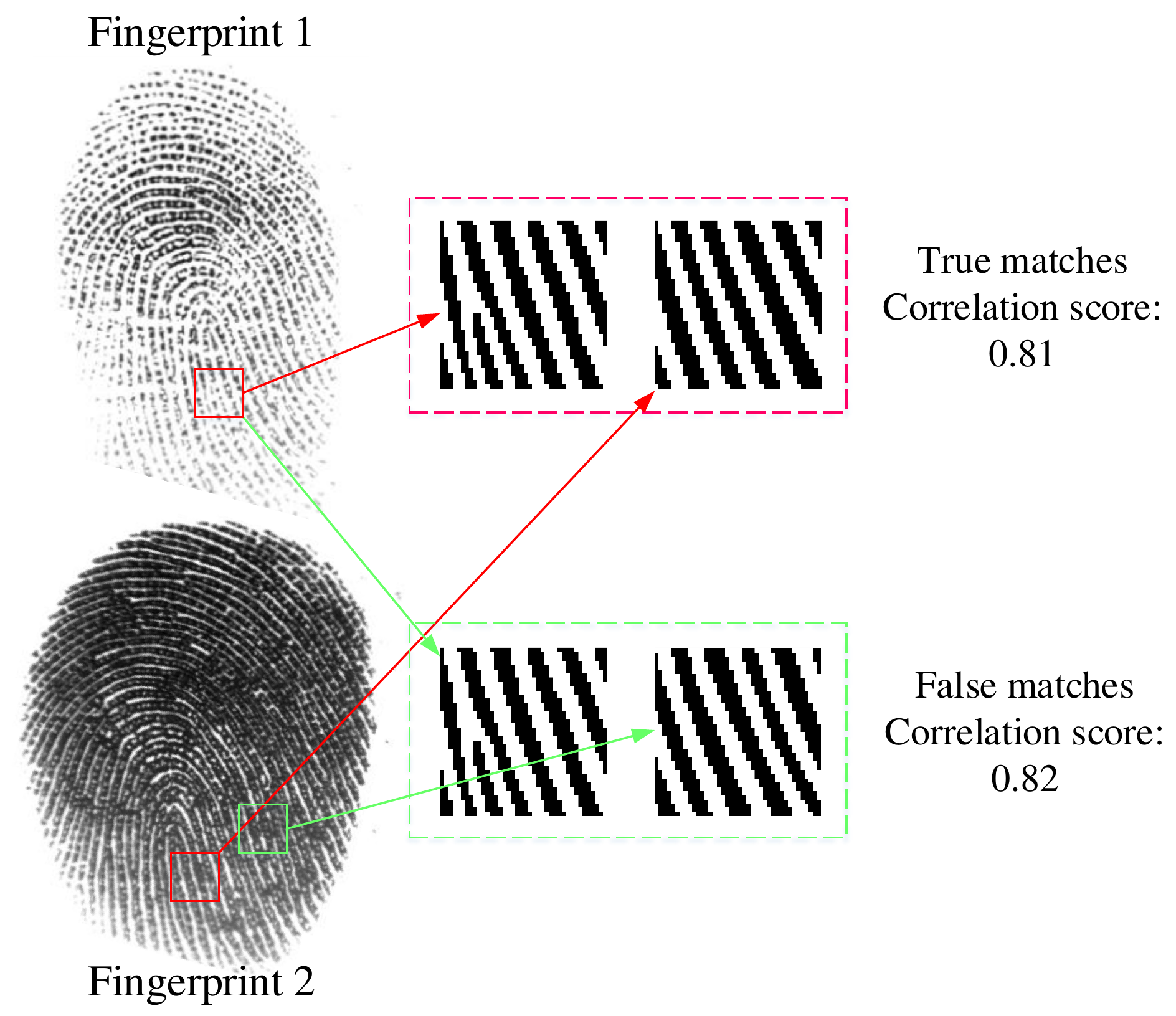}}
  \centerline{(a) Self-Similarity}
\end{minipage}
\hfill
\begin{minipage}[c]{.3\linewidth}
  \centering
  \centerline{\includegraphics[width=\linewidth]{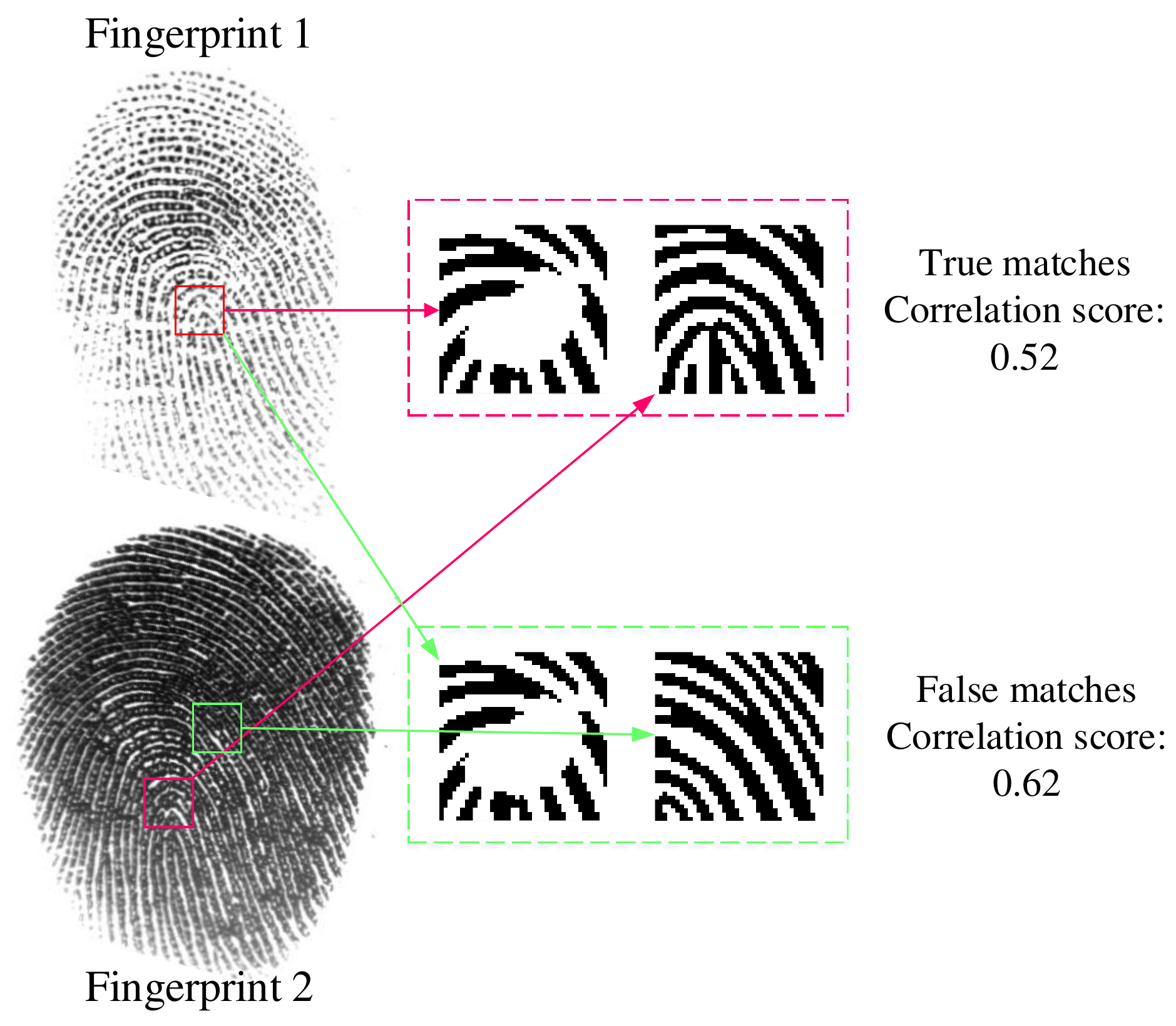}}
  \centerline{(b) Low quality}
\end{minipage}
\hfill
\begin{minipage}[c]{.3\linewidth}
  \centering
  \centerline{\includegraphics[width=\linewidth]{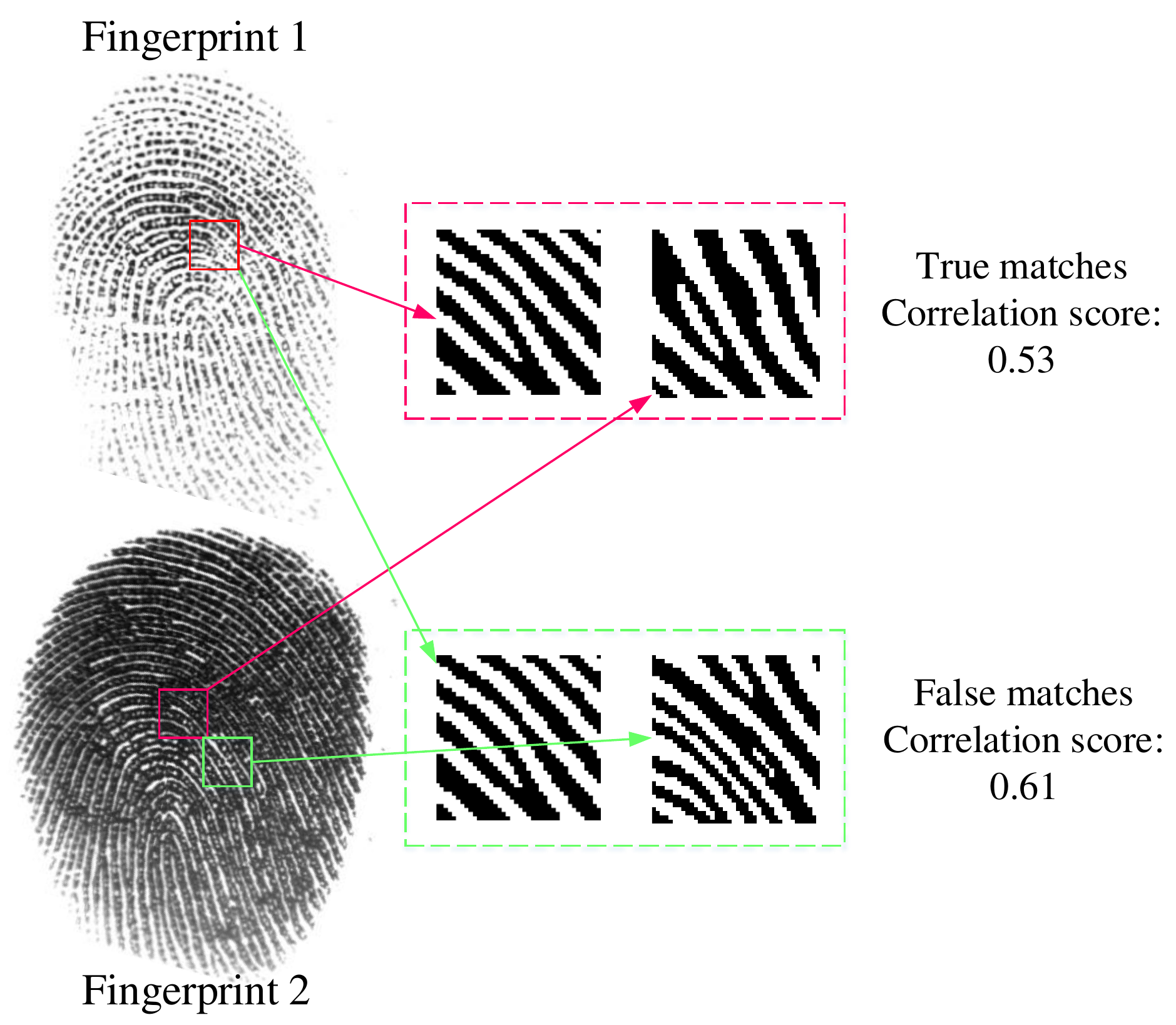}}
  \centerline{(c) Distortion}
\end{minipage}
\caption{Dense registration faces three challenges: high self-similarity of ridge pattern, low image quality, and large distortion. The figure shows three examples where true matching scores are lower than false matching scores according to image correlation coefficient. (a) Because of the self-similarity of fingerprints, the correlation score between true mates may be lower than false mates. (b) Ridge curves differ in low quality regions, which troubles local matching by increasing intra-class variations. (c) Distortion increases intra-class variations and enlarges search space for possible matches.}
\label{fig:challenge}
\end{figure*}

In contrast to minutiae-based methods, dense fingerprint registration \cite{Si2017} aims to obtain pixel-wise displacement measures between two fingerprints with nonlinear skin distortion, instead of producing only sparse displacement measures. The core of dense registration is a local matching problem: for a pixel in the reference fingerprint, find the corresponding pixel in the input fingerprint. Because of self-similarity of ridge pattern, noise, and distortion, local matching is challenged by large intra-class variations among mated regions and small inter-class variations among non-mated regions, as well as large search space. As shown in Fig. \ref{fig:challenge}, noise leads to changes in ridge patterns, and increases intra-class difference; self-similarity makes fingerprint have low inter-class variation, thus difficult to find the true mate among many highly similar non-mated regions; distortion changes ridge patterns, raises intra-class difference, and enlarges search space. 

Regarding these challenges, current dense registration methods \cite{Si2017}\cite{Cui2018} still need improvements from aspects of local displacement estimation and global deformation constraints. The phase demodulation method \cite{Cui2018} is weak in local displacement estimation, as it can only obtain displacement perpendicular to ridge orientation, less than one ridge period, and it is sensitive to noise. The image correlation method \cite{Si2017} is slightly better than phase demodulation method for local displacement estimation. However, due to three aforementioned challenges in dense fingerprint registration, they cannot accurately measure the displacements. Therefore, they all rely heavily on global deformation constraints. The global deformation constraint for phase demodulation is dependent on the order of phase unwrapping, which has the problem of error accumulation \cite{1998unwrap}. The image correlation method uses Markov Random Field (MRF) to implement global deformation constraints, but it cannot solve the problem of large local measurement errors. These drawbacks of existing dense registration methods imply for needs for a more powerful method to overcome those challenges.

\begin{figure*}[!]
\centering
\centerline{\includegraphics[width=\linewidth]{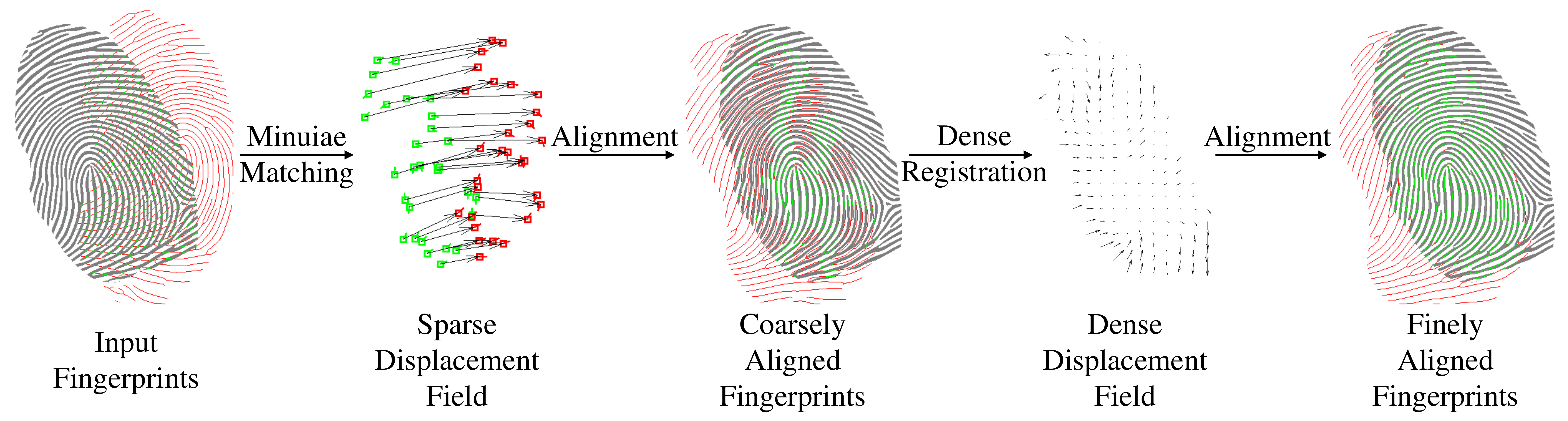}}
\caption{The whole fingerprint registration algorithm consists of two steps: minutiae-based coarse registration and CNN-based fine registration. Note that the proposed dense registration network estimates a pixel-wise displacement field, but for clarity, a block-wise displacement field is shown here.}
\label{fig:flowchart}
\end{figure*}

In recent years, convolutional neural networks prove to provide a better solution to deal with complicated fingerprint recognition problems than handcrafted methods, and has been applied to orientation field estimation \cite{cao2015latent}\cite{schuch2017deep}, minutiae extraction \cite{fingernet2017}, rectification of distorted fingerprints \cite{dabouei2018fingerprint}, fingerprint matching \cite{cao2018automated}, and fingerprint unwarping \cite{Dabouei2019}.

To our knowledge, however, there is no deep learning based method for dense registration of two fingerprints. In this paper, we make an attempt to develop an end-to-end convolutional neural network (CNN) for dense registration of fingerprints, in order to deal with the challenges above. Fig. \ref{fig:flowchart} shows the framework of the whole registration algorithm, which consists of minutiae-based coarse registration and CNN-based fine registration. The input fingerprints are first aligned by fitting a spatial transformation model based on matching minutiae pairs, then go through the proposed dense registration network to get a dense displacement field. The proposed network consists of two parts. The first part is a parallel feature network extracting features from two input fingerprints. The second part is a standard encoder-decoder network \cite{unet2015} to generate displacement field. 

We apply our fingerprint registration algorithm to fingerprint mosaicking, that is, stitching multiple images from one finger into a larger fingerprint image \cite{jain2002fingerprint}. The performance of fingerprint mosaicking relies closely on the accuracy of fingerprint registration. If fingerprint registration fails to align ridge patterns of two fingerprints, fingerprint mosaicking will generate false minutiae and missing minutiae in overlapping region. Based on the results of the  proposed registration algorithm, we proposed an algorithm for selecting the optimal stitching seam to further improve the quality of mosaicking region.

We evaluate our method by comparing with previous dense registration methods on registration accuracy and matching performances. We run experiments on FVC2004 \cite{fvc2004}, Tsinghua Distorted Fingerprint (TDF) database \cite{Si2017} and NIST SD27 latent fingerprint database. Experimental results show that our method outperforms previous dense registration methods in accuracy and is faster. The mosaicking method is evaluated according to consistency in minutiae before and after mosaicking which further demonstrates the performance of our dense registration algorithm.

A previous work of this paper has been published as a conference paper \cite{Cui2019}. As a following work of \cite{Cui2019}, we have made four main improvements:
\begin{itemize}
\item \textbf{Network structure.} The network in \cite{Cui2019} is a local matching network that outputs a single displacement vector from two input fingerprint patches. We upgrade the network to an encoder-decoder structure with smoothing loss that directly outputs a dense displacement field. The new network captures global information and penalizes complex displacement field, which is an important advantage over \cite{Cui2019} to distinguish genuine matching from impostor matching.
\item \textbf{Generation of training data.} The training data in \cite{Cui2019} are selected from FVC2002, which lack distortion and image quality challenges. Also, the training data in \cite{Cui2019} consist of two fingerprint local patches and a corresponding displacement vector, which are of small diversity and lack information of larger scale. In this study, the training data are more plentiful and challenging. The displacement fields are extracted from Tsinghua Distorted Fingerprint Video Database \cite{Si2015} to cover various types of fingerprint distortion. The displacement fields are applied on latent and wet fingerprints to generate data of bad image quality and small area.
\item \textbf{Mosaicking algorithm.} We also develop a fingerprint mosaicking method, which is useful by itself and can reflect the performance of dense registration, and it is not covered in \cite{Cui2019}.
\item \textbf{Experiment.} More experiments are conducted to detailedly examine how and why deep learning method exceeds previous dense registration methods, such as registration result on different kinds of fingerprints, matching performance on latent fingerprints, and mosaicking accuracy for the purpose of examining registration accuracy.
\end{itemize}

The rest parts of this paper are organized as follows. Section \ref{sec:related work} reviews related work. Section \ref{sec:method} proposes our method using end-to-end network. Section \ref{sec:mosaicking} mainly introduces our fingerprint mosaicking method after dense registration. Section \ref{sec:experiment} presents the experimental results of the proposed method versus previous dense registration methods. Finally, Section \ref{sec:conclusion} summarizes our work.

\section{Related Work}\label{sec:related work}
\subsection{Elastic Registration of Fingerprints}\label{ssec:fingerprint_registration}
Early fingerprint registration methods use rigid model to transform fingerprints \cite{Tico2003}, which cannot align minutiae accurately, not to speak of aligning ridges. Thin-plate spline (TPS) model is introduced by several researchers to deal with elastic fingerprint distortion \cite{Bazen2003}\cite{Ross2005} to solve this problem.

TPS based algorithms mainly use minutiae correspondences to fit transformation models, thus the accuracy of these methods relies on the accuracy of minutiae matching. Although lots of work focus on improving minutiae matching accuracy \cite{Tico2003}\cite{Cappelli2010}\cite{cheng2013minutiae}\cite{paulino2012latent}, the performance of minutiae matching is still limited by distortion and bad image quality. Additionally, minutiae-based matching only provides minutiae pairs, and the fitted TPS model cannot align fingerprint area lacking matching minutiae.

Several methods \cite{ross2005fingerprint}\cite{choi2007fingerprint}\cite{choi2011fingerprint}\cite{lan2019non} further incorporate ridge skeleton features correspondences to register fingerprints. However, these methods are sensitive to ridge skeleton errors, which is very common in bad quality fingerprints. They also lack reasonable constraint for deformation and strong descriptors for establishing accurate correspondences between ridge points which are featureless.

\subsection{Dense Registration of Fingerprints}\label{ssec:dense_registration}
Dense registration of fingerprints aims to provide pixel level correspondences between two fingerprints, rather than very sparse minutiae correspondences.

Dense registration of fingerprints is first introduced by \cite{Si2017}. Their method uses image correlation on blocks as similarity measurement, then minimizes the global similarity score to find dense correspondences between two fingerprints. Their method performs better than minutiae-based method, but the computation of image correlation is sensitive to those challenges mentioned in Fig. \ref{fig:challenge} and is computationally expensive.

Phase demodulation method proposed in \cite{Cui2018} transfers the concept of phase demodulation in communication into fingerprint registration. They make use of the fingerprint characteristic that fingerprint patterns look like a 2-D cosine wave, therefore, computing phase shift is equal to computing displacement. But the phase feature they utilized in this method is still a handcraft feature extracted from fingerprint ridge images, and the accuracy of phase feature is bothered in low quality and large distortion area.
%
%

\section{Fingerprint Registration}\label{sec:method}
Our dense registration algorithm mainly consists of three steps: (1) initial registration (or minutiae based fingerprint registration) which finds out the mated minutiae pairs from two fingerprints and roughly aligns them by minutiae correspondences; (2) fine registration using an end-to-end network which outputs a dense displacement field from roughly aligned fingerprints; (3) dense registration of the input fingerprint to the reference fingerprint according to the result of step (2). 
\subsection{Initial Registration}\label{ssec:initial_registration}
This step performs a coarse registration based on matching minutiae pairs, which is also used in previous dense registration methods \cite{Si2017}\cite{Cui2018}. In this step, we first use VeriFinger SDK \cite{verifinger} to extract minutiae, and compute similarity scores among all minutiae pairs using MCC minutia descriptor \cite{Cappelli2010}. Then, we use spectral clustering method \cite{Leordeanu2005} to find those most probable matching minutiae pairs. Finally, those matching minutiae are used as landmark points to compute a thin plate spline (TPS) model \cite{Bookstein1989}, which is then used to align the input fingerprint to the reference fingerprint.

\begin{figure*}[!]
\centering
\centerline{\includegraphics[width=\linewidth]{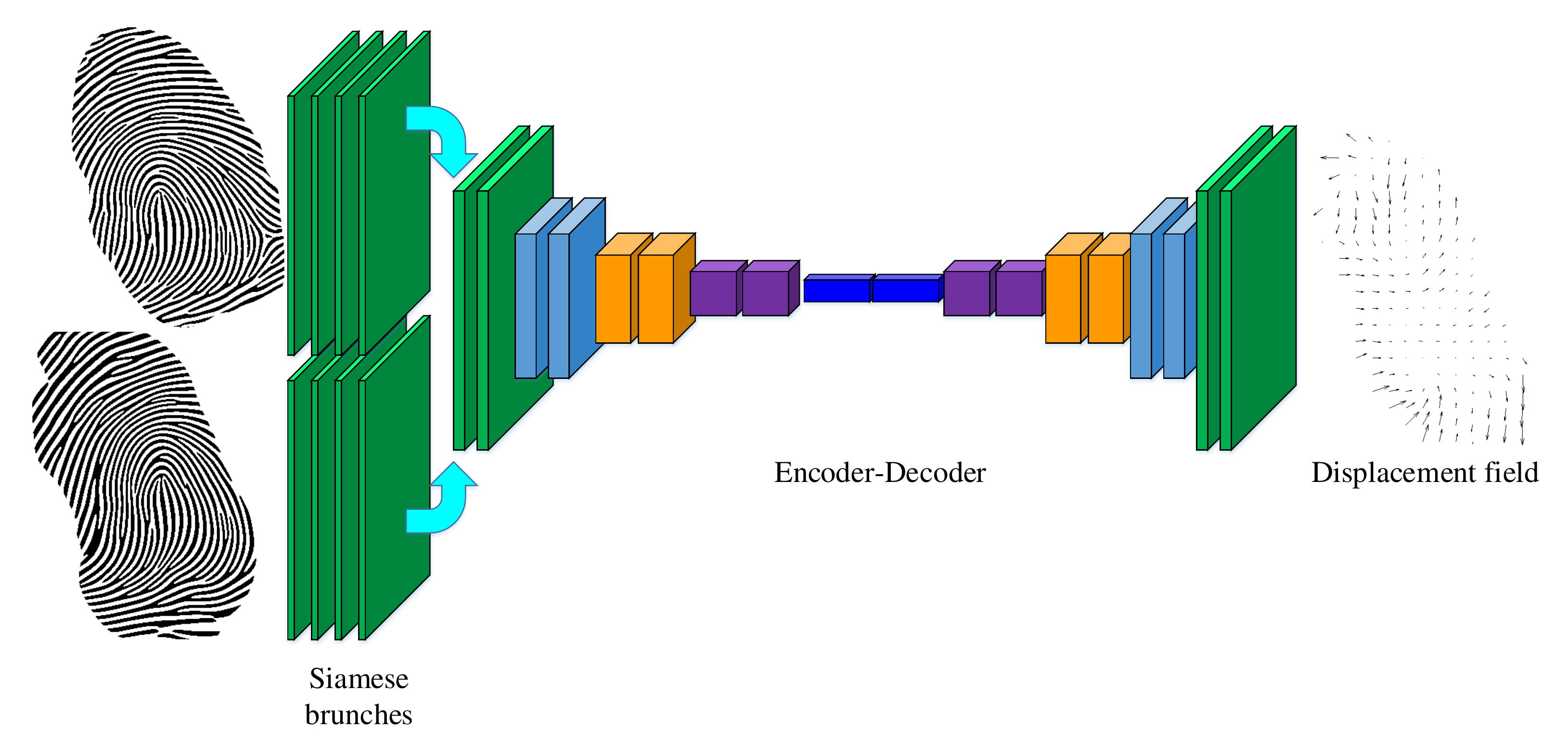}}
\caption{Architecture of the proposed network, including a siamese network for feature embedding and an encoder-decoder for estimating displacement field. Siamese brunches consist of 4 convolution layers. Encoder-decoder includes 5 down layers and 4 up layers. Layers in different colors imply for convolution layers in different sizes from maxpooling or unsampling.}
\label{fig:network}
\end{figure*} 

\subsection{Network for Displacement Field Estimation}\label{ssec:initial_registration}
As shown by Fig. \ref{fig:network}, the proposed network first uses two parallel brunches to extract features from two input fingerprints. Then the two features are concatenated and sent into an encoder-decoder to regress the displacement field $\mathcal{D}=(\mathcal{D}_x,\mathcal{D}_y)$ at each pixel. The network is fully convolutional, therefore can be trained and evaluated in an end-to-end manner, and can deal with fingerprint images of arbitrary size.

The training loss consists of two parts: the regression loss between estimated displacement field $\mathcal{D}_{est}$ and ground-truth displacement field $\mathcal{D}$, and the smoothing loss of estimated $\mathcal{D}_{est}$.
\begin{equation}
\mathcal{L}=\mathcal{L}_{est}+\lambda\mathcal{L}_{smo}
\end{equation}
\begin{equation}
\mathcal{L}_{est}=\sum_{}\left \| \mathcal{D}_{est}(x)-\mathcal{D}(x) \right \|_2^2
\end{equation}
\begin{equation}
\mathcal{L}_{smo}=\sum_{x}\left \| \nabla(\mathcal{D}_{est}(x)) \right \|_2^2
\end{equation}
where $x$ denotes a pixel and $\lambda$ is empirically set to 0.8. 

\begin{figure*}[htb]
\centering
\centerline{\includegraphics[width=\linewidth]{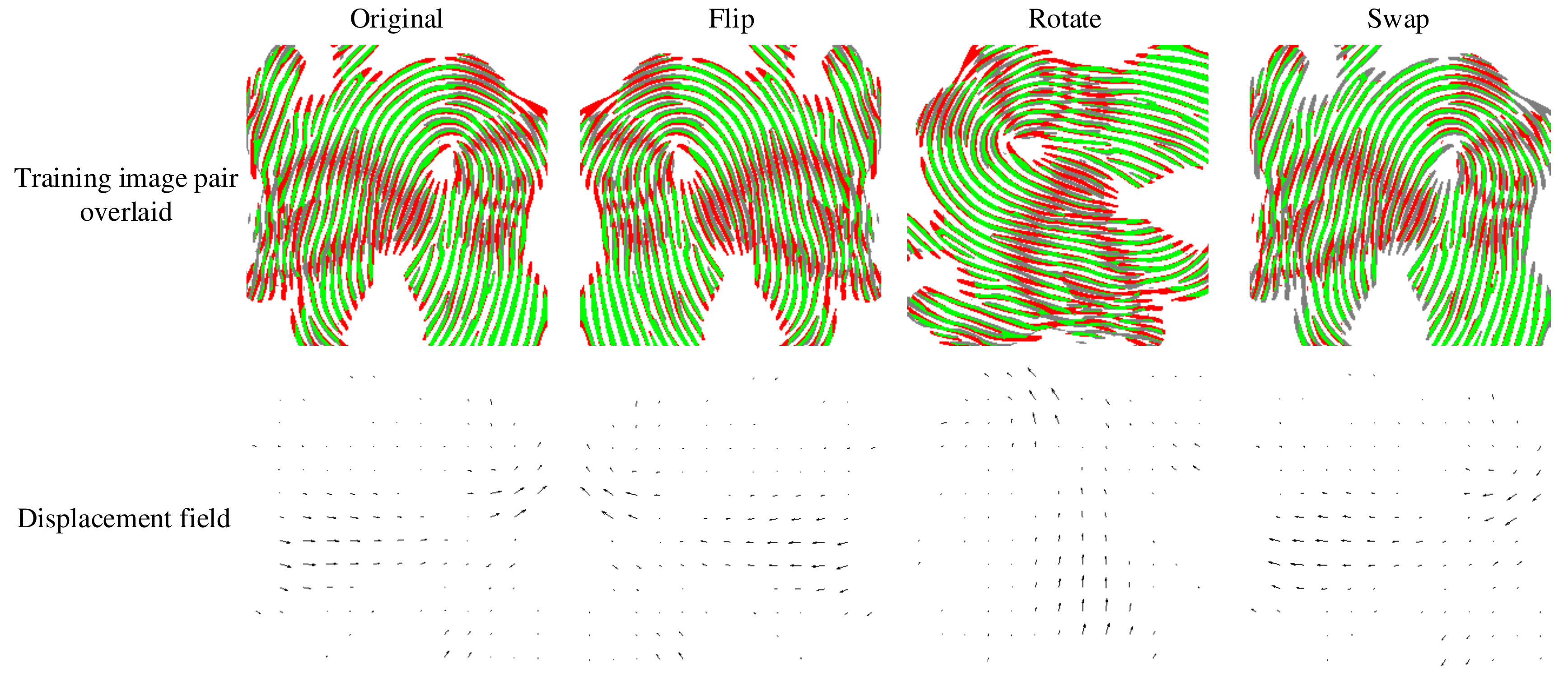}}
\caption{Examples of three data argumentation methods, including flipping, rotating, and swapping. The first row is training image pairs displayed by overlapping two fingerprint images. Binarized input fingerprints are overlaid on the binarized reference fingerprint (gray line) for visualization. Green pixels indicate regions where input fingerprint ridges align with reference fingerprint ridges, and red pixels indicate non-overlapping regions between input and reference fingerprint ridges. The second row is displacement fields displayed as vector fields.}
\label{fig:argument}
\end{figure*} 

\subsection{Training Data}\label{ssec:train_data}
For deep learning studies, training data is as important as network structure. To train an end-to-end network in Fig. \ref{fig:network}, we need sufficient pairs of mated fingerprints and corresponding dense displacement filed. However, acquiring a large size of ground truth data of displacement fields from challenging mated fingerprints manually is a gigantic task, and is unrealistic.

Therefore, we use Tsinghua Distorted Fingerprint Video Database \cite{Si2015} to generate displacement fields. We select this database for two reasons: 1) A large proportion of fingerprints in this database are of severe distortion, which is beneficial for training a network suitable for large distortion. 2) Although large distortion may cause trouble in fingerprint matching, fingerprint images in adjacent frames are of little translation. By tracing the motion of minutiae between neighboring frame one after another, we can construct reliable minutiae correspondences across all frames even if they are highly distorted. Therefore, we can get displacement field using minutiae pairs to compute a TPS transform. 

As a result, we obtain 320 displacement fields altogether from Tsinghua Distorted Fingerprint Video Database. But we do not directly use these data to train a network because 320 pairs of fingerprints of good image quality are clearly not enough from aspects of fingerprint image quality and fingerprint patterns. We build training data from three datasets listed in Table \ref{table:datatrain}, i.e. rolled fingerprints, latent fingerprints, and wet fingerprints. Latent fingerprints are provided by local police department, and they are used in order to make our method insensitive to various latent fingerprint qualities. Wet fingerprints were also used to improve our method on wet fingerprint circumstance which is not included in latent fingerprints. 

\newcommand{\tabincell}[2]{\begin{tabular}{@{}#1@{}}#2\end{tabular}}
\begin{table*}[htb]
\caption{Fingerprint databases used as training data.}  
\centering
\begin{tabular}{|c| c |c|} 
\hline
Databases & Description & Training data \\
\hline 
Tsinghua distorted fingerprint video & largely distorted fingerprints & 320 pairs of displacement fields\\
\hline 
Rolled fingerprint & 10,459 rolled fingerprints from local police&$10,459\times 2=20,918$ pairs of fingerprints \\
\hline 
Latent fingerprint & 10,459 latent fingerprints from local police& $10,459\times 2=20,918$ pairs of fingerprints\\
\hline 
Wet fingerprint & 30 wet fingerprints from FVC2004 DB1/2/3{\_}B & $30\times 320=9,600$ pairs of fingerprints\\
\hline 
\end{tabular}
\label{table:datatrain}
\end{table*}

For a fingerprint image $I_i$ in dataset, we use displacement field $\mathcal{D}_j$ to transform and interpolate a transformed fingerprint image $I_{ij}=\mathcal{D}_j(I_i)$. A pair of images $I_i$ and $\mathcal{D}_j(I_i)$ are first binarized by VeriFinger \cite{verifinger} to get enhanced images, then we crop areas of size $256{\times}256$ from enhanced images and displacement field. Thereby a set of training data \{$I_1, I_2, \mathcal{D}$\} is generated.

For latent fingerprints, they are additionally enhanced by FingerNet \cite{fingernet2017} for better results. We have 10,459 latent fingerprint images and 320 displacement fields, and each latent fingerprint image is combined with 2 random selected displacement fields. Therefore, a total number of 20,918 pairs of training dataset are generated. We do not combine each fingerprint image with each displacement field to reduce the redundancy of training data. For wet fingerprints, because we only have 30 wet fingerprints from FVC2004 DB1/2/3{\_}B, each wet fingerprint is combined with all 320 displacement fields.

We conduct data augmentation by mirror flipping images, rotating, and swapping the position of $I_1$ and $I_2$, as shown in Fig. \ref{fig:argument}. When changing fingerprint image patches, their displacements are changed at the same time. The displacement field $\mathcal{D}$ here has size $256{\times}256{\times}2$ which is the same as input images. Therefore, flipping and rotating are also operated on $\mathcal{D}$. For clearer explanation, we split $\mathcal{D}$ into $\mathcal{D}_x$ and $\mathcal{D}_y$ in following explanation. 

The three augmentation types are:
\begin{itemize}
  \item [1)] \textbf{Mirror flipping:} Augmented \{$I_1, I_2, \mathcal{D}_x, \mathcal{D}_y$\} at coordinate $(i,j)$ = Original \{$I_1, I_2, -\mathcal{D}_x, \mathcal{D}_y$\} at coordinate $(-i,j)$, where axis origin is located at the center of image.
  \item [2)] \textbf{Rotate:} Augmented \{$I_1, I_2, \mathcal{D}_x, \mathcal{D}_y$\} at coordinate $(i,j)$ = Original \{$I_1, I_2, \mathcal{D}_x\cos\theta+\mathcal{D}_y\sin\theta, -\mathcal{D}_x\sin\theta+\mathcal{D}_y\cos\theta$\} at coordinate $(i\cos\theta+j\sin\theta,-i\sin\theta+j\cos\theta)$, $\theta\in[90^\circ,180^\circ,270^\circ]$, where axis origin is located at the center of image.
  \item [3)] \textbf{Swap:} Augmented \{$I_1, I_2, \mathcal{D}_x, \mathcal{D}_y$\} at coordinate $(i,j)$ = Original \{$I_2, I_1, -\mathcal{D}_x, -\mathcal{D}_y$\} at coordinate $(i,j)$, where axis origin is located at the center of image.
\end{itemize}

The three augmentation types can be integrated to enlarge the size of training data. After augmentation, the training set is 16 times bigger, which meets the need of training our network.

\subsection{Dense Registration}\label{ssec:interp}
The final registration result is a direct nearest neighbor interpolation using the output displacement $\mathcal{D}$ to generate an aligned input fingerprint. Previous dense registration methods \cite{Si2017}\cite{Cui2018}\cite{Cui2019} use TPS transform to conduct the final step of dense registration because these methods output displacement fields on grid level, size of $20 \times 20$ pixels for instance. Therefore, fitting a TPS transform is needed to calculate displacements on the rest of pixels. 

Our method outputs displacement field directly on pixel level, and the output displacement field is already smoothed because of smoothing loss in training network. So nearest neighbor interpolation can be used in the final step. Comparing with TPS interpolation, nearest neighbor interpolation is much faster.

\begin{figure*}[htb]
\centering
\centerline{\includegraphics[width=\linewidth]{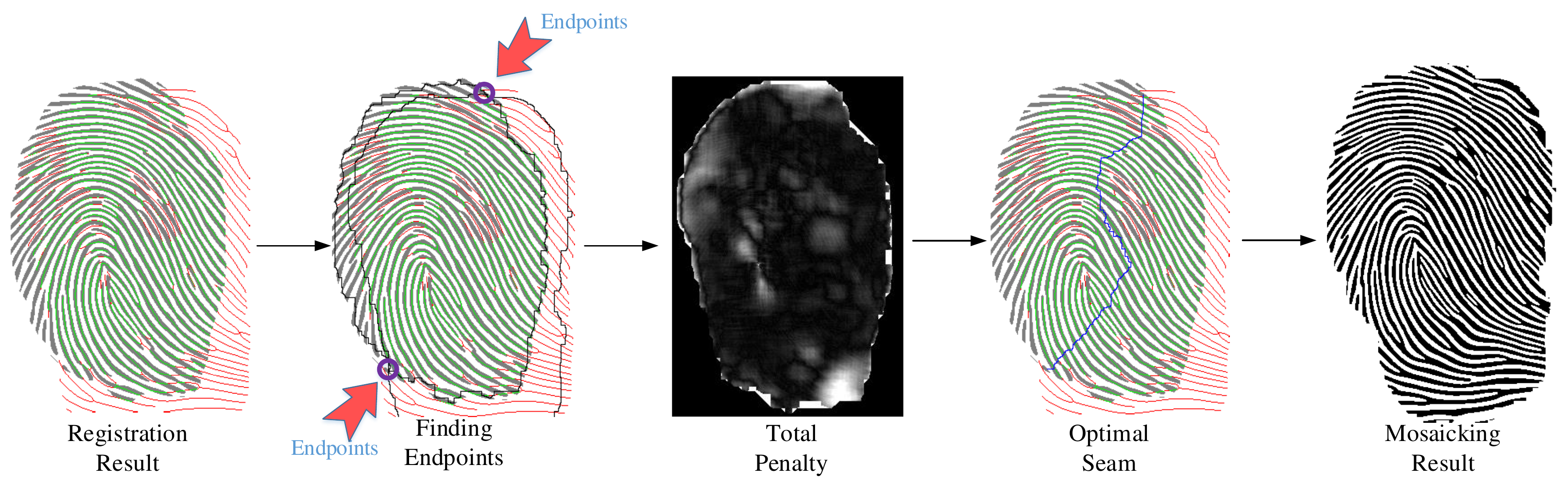}}
\caption{The flowchart of the proposed fingerprint mosaicking algorithm.}
\label{fig:mosaicking}
\end{figure*} 

\section{Fingerprint Mosaicking}\label{sec:mosaicking}
An important application of fingerprint registration is fingerprint mosaicking, which combines a series of flat fingerprints to reconstruct a full-size rolled fingerprint. Comparing with conventional fingerprint mosaicking method that combines a list of rolled fingerprints \cite{ratha1998image}, the proposed fingerprint mosaicking method aims at stitching multiple fingerprints from different acquisition sessions, which is more general and complicated. To combine multiple flat fingerprints into one fingerprint, these flat fingerprints need to be pre-aligned, and the performances of registration will consequently affect the performances of mosaicking. 

In this section, a fingerprint mosaicking method is introduced after applying our registration algorithm. A pair of fingerprints may still have some differences after registration, and a straightforward method based on weighted fusion of fingerprints will lead to mess in badly aligned region \cite{choi2005fingerprint}. To overcome this problem, we develop an optimal seam selection method to combine two fingerprints and reduce influence from badly registered area. The mosaicking algorithm is detailedly introduced in next subsections. \ref{ssec:mosaicking_method} introduces the proposed mosaicking method, and \ref{ssec:multi_mosaicking} talks about the utilization of the proposed mosaicking method on fusing multiple fingerprint images.

\subsection{Mosaicking of Two Fingerprints}\label{ssec:mosaicking_method}
As Fig. \ref{fig:mosaicking} illustrates, our fingerprint mosaicking method consists of three steps: 1) find endpoints of the seam to be determined, 2) build an undirected graph based on the differences between two fingerprints, and 3) find the optimal seam and conduct mosaicking. 

Step 1 is quite straightforward: the intersection points of the edges of the two fingerprints are considered as endpoint answers, as shown in Fig. \ref{fig:mosaicking}. Step 2 builds an undirected graph using each pixel in the overlapping region as a node, and its 4-connection neighbors as edges. The edges are weighted by the penalties computed from the differences between two fingerprints.

The differences of two fingerprints, also viewed as the penalty of separation seam, contains three parts: the gray-scale image difference $\left| I_1(p)-I_2(p)\right|$, the orientation difference $\left| O_1(p)-O_2(p)\right|$ between reference fingerprint and registered input fingerprint, and the penalty by the distance to the edges. The first two penalties quantify how much the two fingerprints differ, and the third penalty simply prevents the separation seam from being too close to one side, but instead being in the middle and making use of both fingerprints. Clearly, the separation seam must be within the overlapping region of two fingerprints, thus we only compute penalties in the overlapping region.

\begin{equation}
\begin{aligned}
\mathcal{P}\left(p\right)&=\left| I_1(p)-I_2(p)\right|+\lambda_1\left| O_1(p)-O_2(p)\right|\\
&+\lambda_2 \exp\left(-distance\left( p,edges\right)\right)
\end{aligned}
\label{equ:penalty}
\end{equation}

After settling endpoints of the possible seam and constructing the graph, the optimal seam is determined in step 3. To combine reference fingerprint and aligned input fingerprint into a larger fingerprint image, we need to find an optimal separation line. The area on the left of separation line from input fingerprint, and the area on the right of separation line from reference fingerprint, are stitched to form a full fingerprint image. The optimal separation line, or optimal seam, is defined as minimizing the differences between two sides across this seam. Then aligned input and reference fingerprints are cut according to the selected seam, and they are piecing together to get the mosaicked result.

Equation \ref{equ:penalty} defines a pixel-wise penalty function, and the total penalty of a candidate seam is the sum of penalties of the pixels going through by this seam. Therefore, finding optimal separation line can be viewed as finding an optimal path with minimum total penalty, which is a standard \textit{shortest path} issue and can be solved by \textit{Dijkstra} algorithm. 


\begin{figure*}[!]
\begin{minipage}[b]{1\linewidth}
  \centering
  \centerline{\includegraphics[width=.95\linewidth]{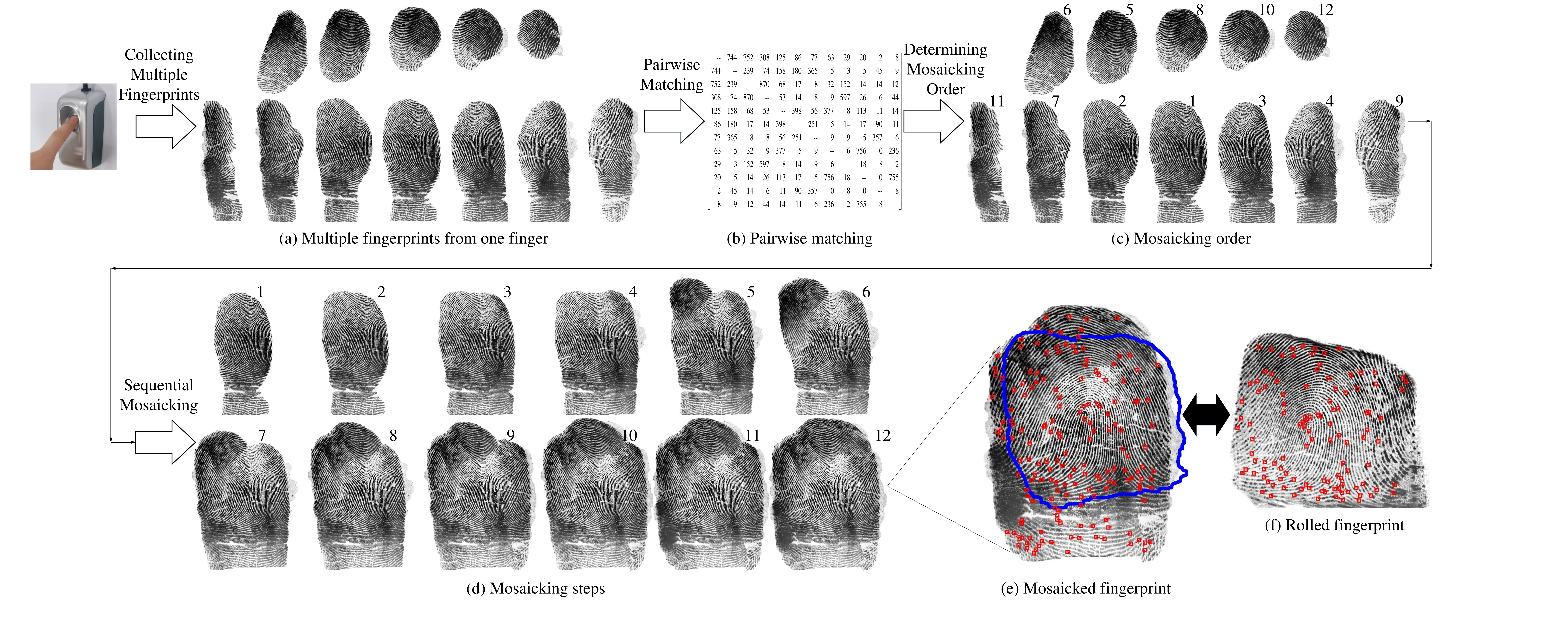}}
\end{minipage}
\caption{Growing process of multiple fingerprint mosaicking. The last two images are a comparison between the mosaicked full fingerprint and a traditional rolled fingerprint, where the blue outline on the mosaicked fingerprint refers to corresponding rolled fingerprint area. Comparing with the mosaicked fingerprint, a single rolled fingerprint cannot get a full fingerprint due to the special shape of finger, and hence contains much fewer minutiae.}
\label{fig:mosaic_images}
\end{figure*}

\subsection{Mosaicking of Multiple Fingerprints}\label{ssec:multi_mosaicking}
The above mosaicking algorithm can be extended to combine a series of flat fingerprints into a full fingerprint. These fingerprints may be gathered by rolling finger on a fingerprint sensor, or just by pressing finger in different angles on the sensor. By pressing finger with various angles, we can get fingerprints from different parts on finger, as shown in Fig. \ref{fig:mosaic_images}. We collect 12 images to cover nearly all areas on finger in order to reconstruct a full fingerprint as large as possible. Because of the special shape of finger, traditional rolled fingerprint cannot cover all fingerprint area, especially fingertip area. Meanwhile, gathered multiple flat fingerprints can fully cover whole fingerprint region. In addition, acquisition of rolled fingerprint requires larger sensors, which are more expensive and much less popular than small area fingerprint sensors.

Before combining fingerprints, we need to know the relations between those fingerprints to determine the order of mosaicking, i.e. which fingerprint should be stitched first. An optimal mosaicking sequence should maximize fingerprint similarities between adjacent fingerprints to reduce probable registration and mosaicking errors. Therefore, we run pairwise fingerprint matching by VeriFinger to compute similarity scores between all fingerprints. The first fingerprint is chosen as the one with  the largest total similarity scores with all other fingerprints. The remaining fingerprints are mosaicked onto the first one sequentially in a descending order of their matching scores to it. 

Because of the special shape of finger, just rolling finger once cannot cover the whole fingerprint area, especially for fingerprint tip region. Mosaicking multiple fingerprints from different acquisition sessions is needed to obtain a full fingerprint. Fig. \ref{fig:mosaic_images} also shows each step of mosaicking multiple fingerprints. A new fingerprint is first registered to the mosaicked result of last step, then they are stitched together to make a larger fingerprint. The last two fingerprint images in Fig. \ref{fig:mosaic_images}(e)(f) show mosaicked full fingerprint and a conventional rolled fingerprint for comparison. Comparing with rolled fingerprint, the mosaicked fingerprint has larger area and more minutiae (169 vs. 114), which is beneficial for fingerprint recognition.

\begin{figure}[!]
\begin{minipage}[b]{1\linewidth}
  \centering
  \centerline{\includegraphics[width=\linewidth]{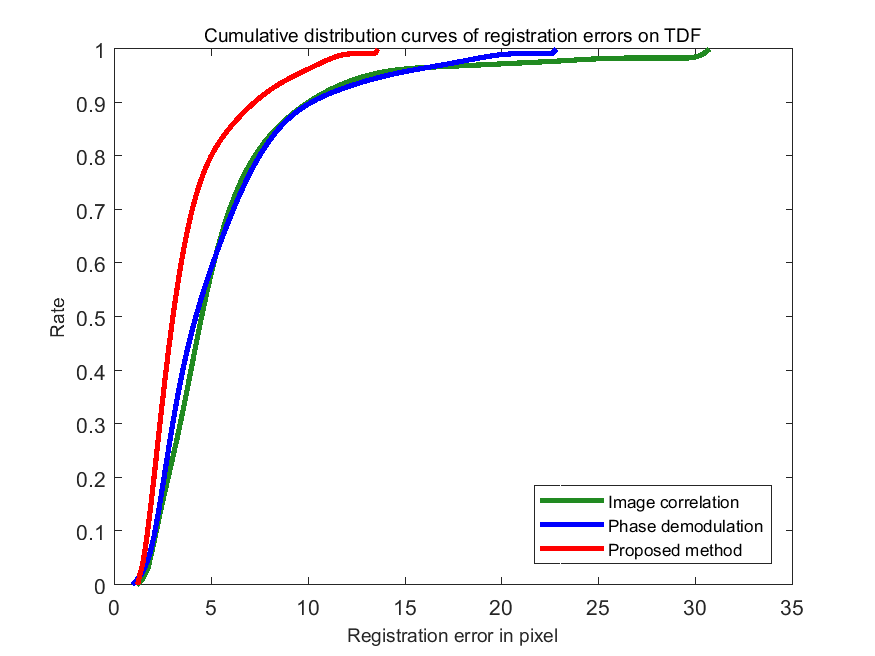}}
\end{minipage}
\caption{Cumulative distribution curves of registration errors on TDF database using image correlation (green), phase demodulation (blue), and the proposed method (red).}
\label{fig:reg_error}
\end{figure}

\begin{figure*}[!]
\begin{minipage}[b]{\linewidth}
  \centering
  \centerline{\includegraphics[width=\linewidth]{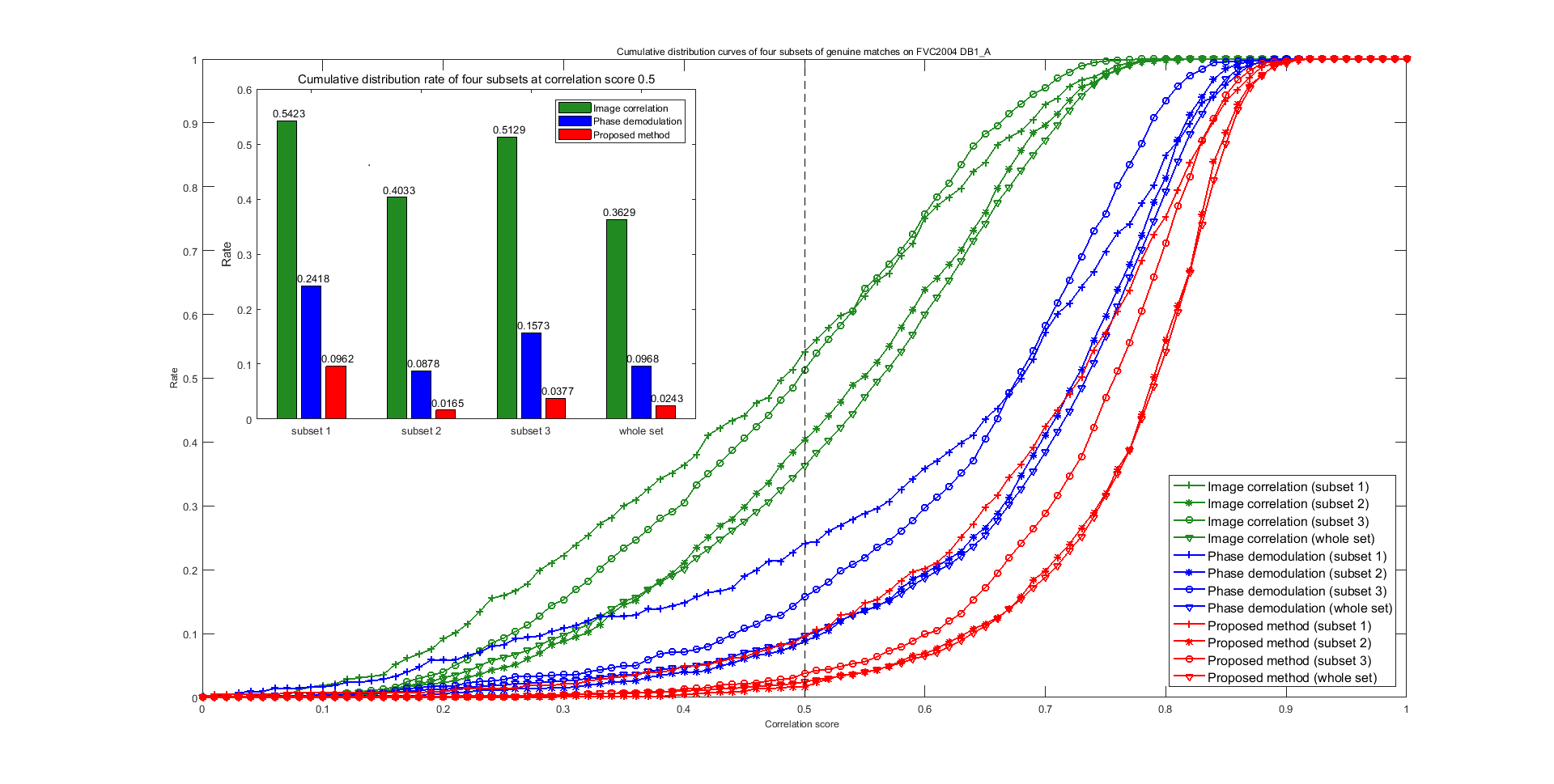}}
\end{minipage}
\caption{Cumulative distribution curves of image correlation scores on four subsets of genuine matching pairs on FVC2004 DB1{\_}A. Matching score is computed as image correlation score between fingerprints aligned by each of the three different dense registration algorithms, using image correlation (green), phase demodulation (blue), and the proposed method (red). The four subsets are manually marked distorted fingerprints, dried fingerprints, moistened fingerprints, as well as all fingerprints for average performance. The subfigure in the upper left area detailedly displays the cumulative probabilities on four subsets by dense registration algorithms at correlation score 0.5.}
\label{fig:fmr_2}
\end{figure*}

\begin{figure}[!]
\begin{minipage}{\linewidth}
  \centering
  \centerline{\includegraphics[width=\linewidth]{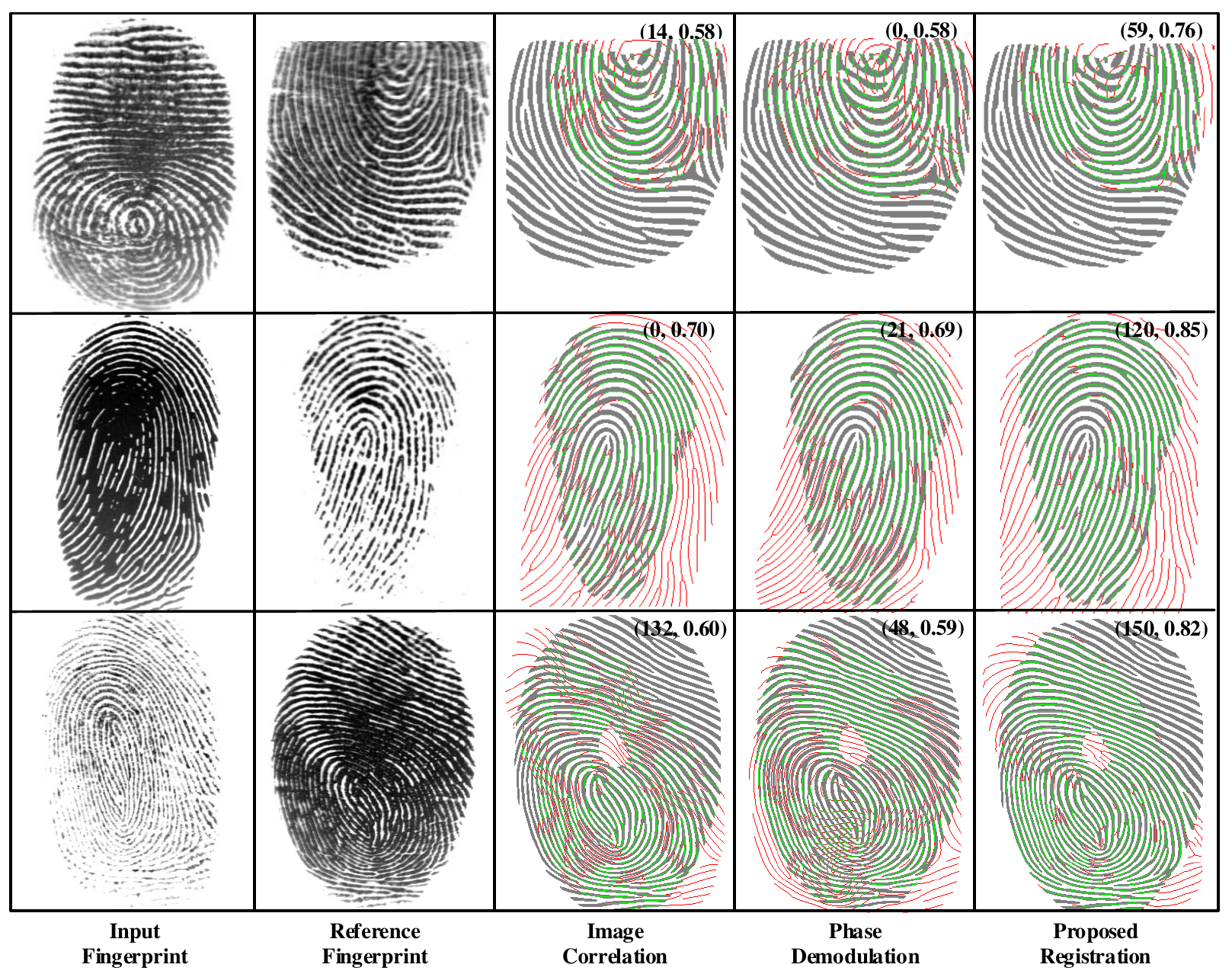}}
\end{minipage}
\caption{Registration examples of different dense registration methods for genuine matching fingerprints. The numbers in the brackets are matching scores by VeriFinger and image correlator.}
\label{fig:reg_examples}
\end{figure}

\begin{figure}[!]
\begin{minipage}[b]{\linewidth}
  \centering
  \centerline{\includegraphics[width=\linewidth]{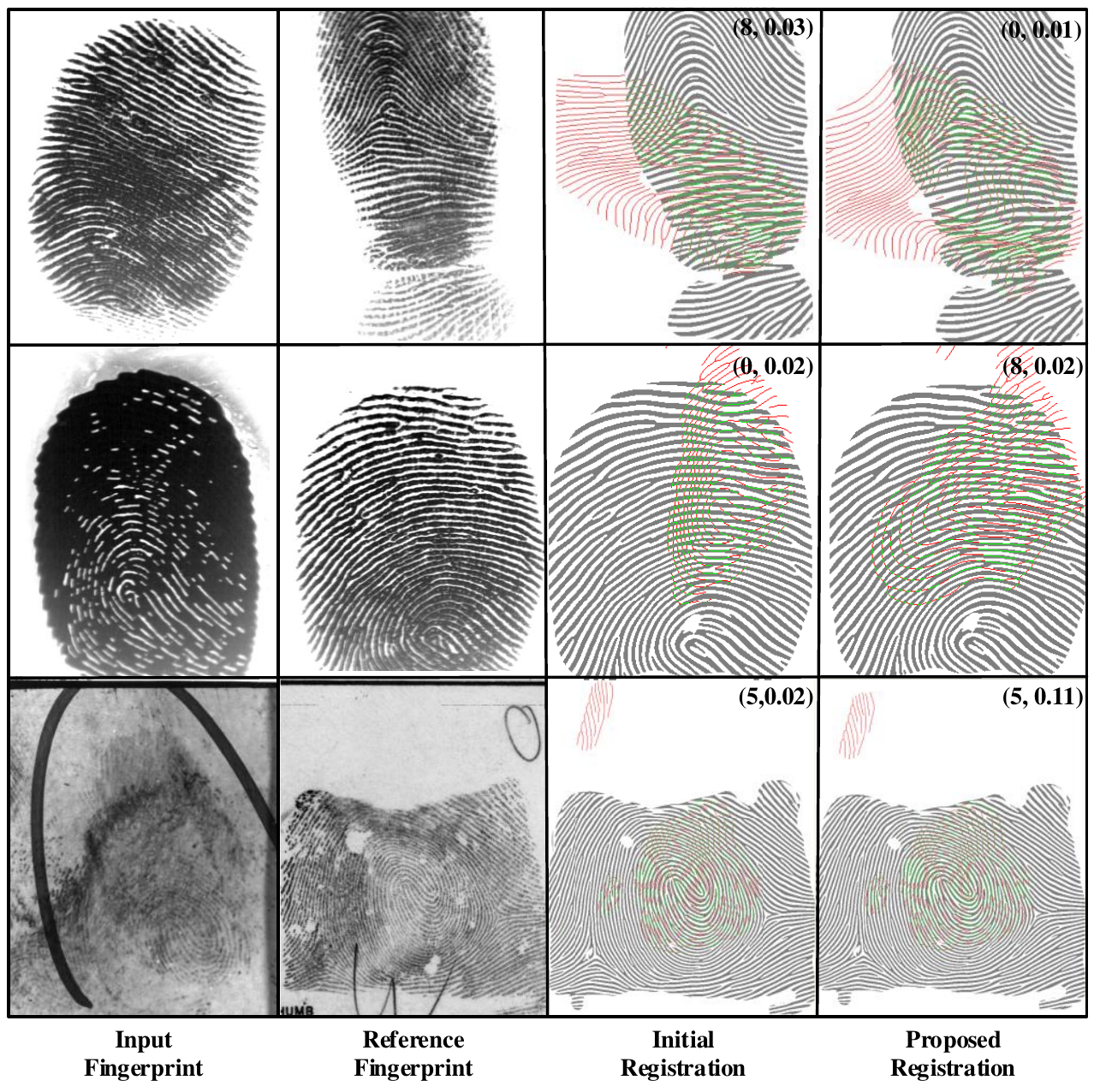}}
\end{minipage}
\caption{Bad registration examples by the proposed dense registration method. Due to distortion and small overlapping region (row 1), bad quality and distortion (row 2), and extremely low quality (row 3), minutiae-based initial registration is far from correct, which is beyond the capability of dense registration method.}
\label{fig:failed_example}
\end{figure}

\begin{figure*}[!]
\begin{minipage}[b]{.49\linewidth}
\centering
\centerline{\includegraphics[width=\linewidth]{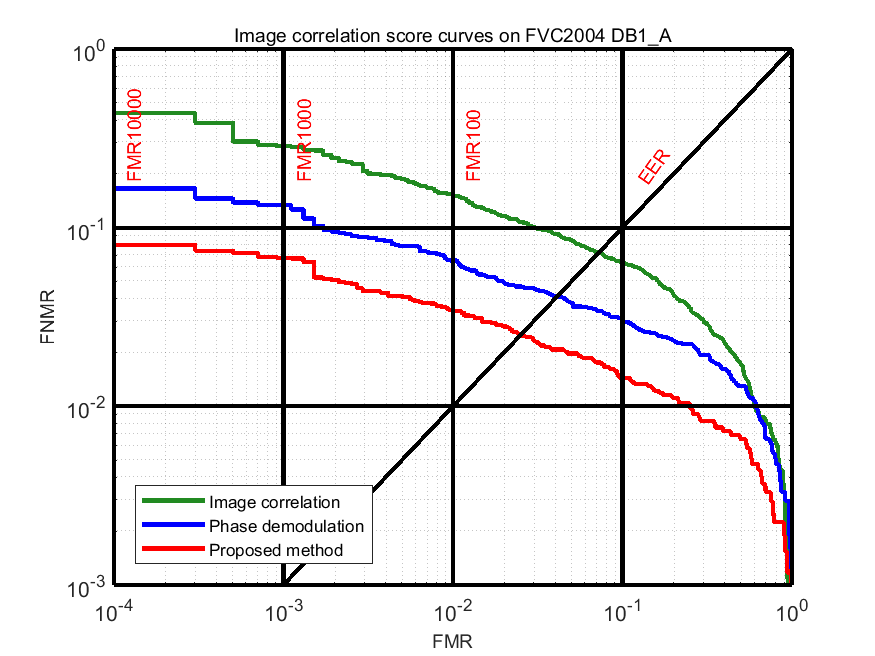}}
\centerline{(a) Image correlator on DB1{\_}A}\medskip
\end{minipage}
\hfill
\begin{minipage}[b]{.49\linewidth}
\centering
\centerline{\includegraphics[width=\linewidth]{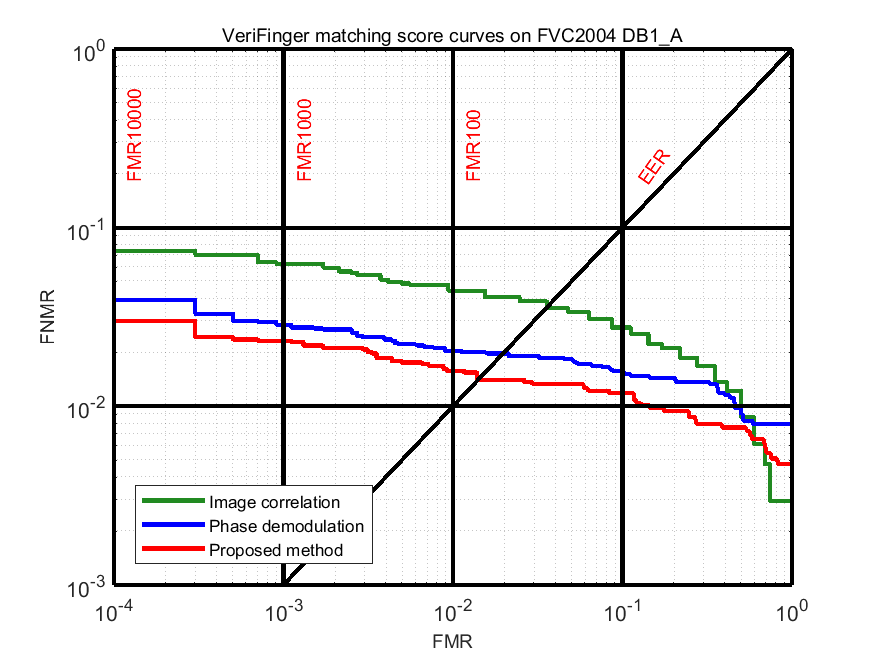}}
\centerline{(b) VeriFinger matcher on DB1{\_}A}\medskip
\end{minipage}
\hfill
\begin{minipage}[b]{.49\linewidth}
\centering
\centerline{\includegraphics[width=\linewidth]{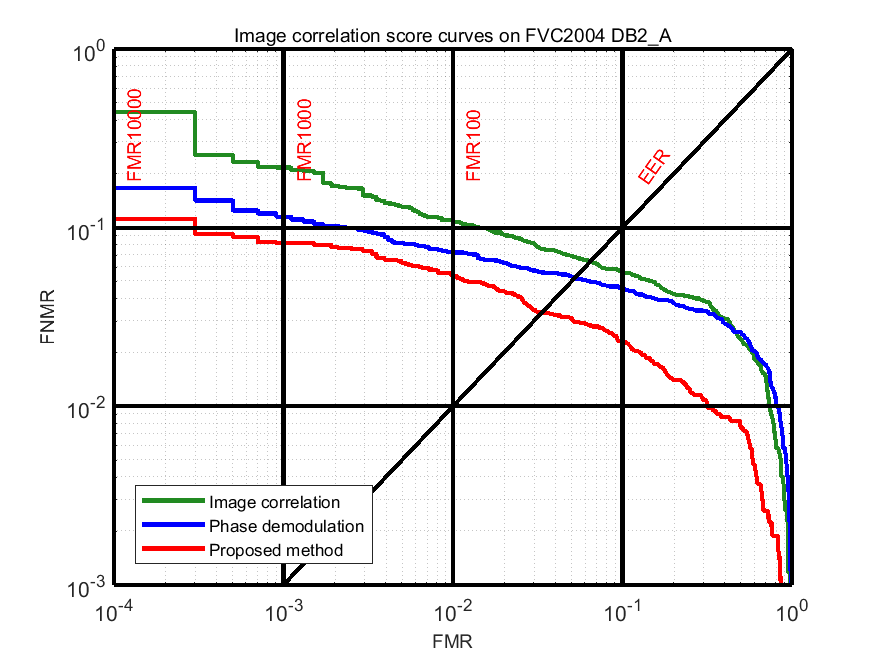}} 
\centerline{(c) Image correlator on DB2{\_}A}\medskip
\end{minipage}
\hfill
\begin{minipage}[b]{.49\linewidth}
\centering
\centerline{\includegraphics[width=\linewidth]{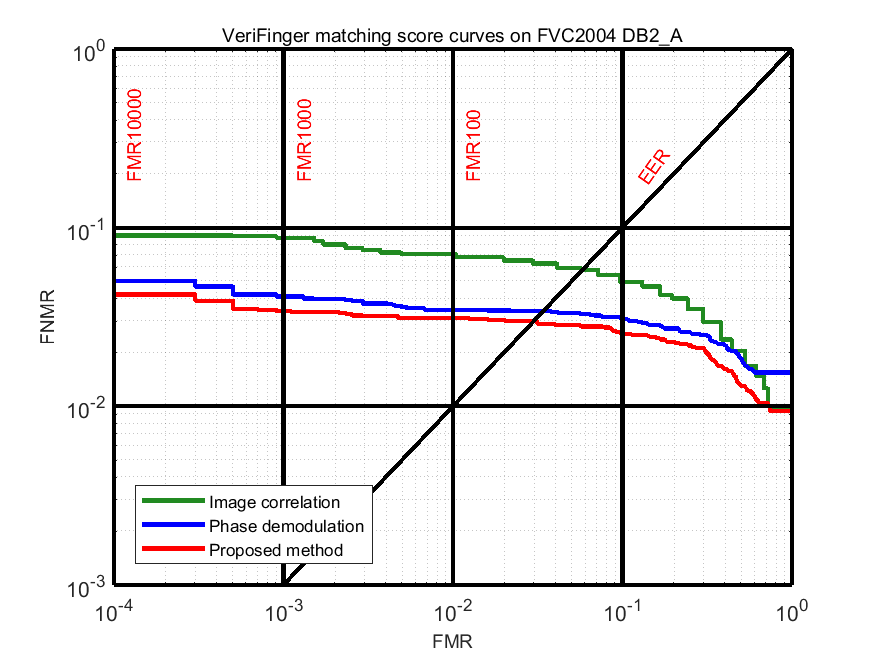}}
\centerline{(d) VeriFinger matcher on DB2{\_}A}\medskip
\end{minipage}
\hfill
\begin{minipage}[b]{.49\linewidth}
\centering
\centerline{\includegraphics[width=\linewidth]{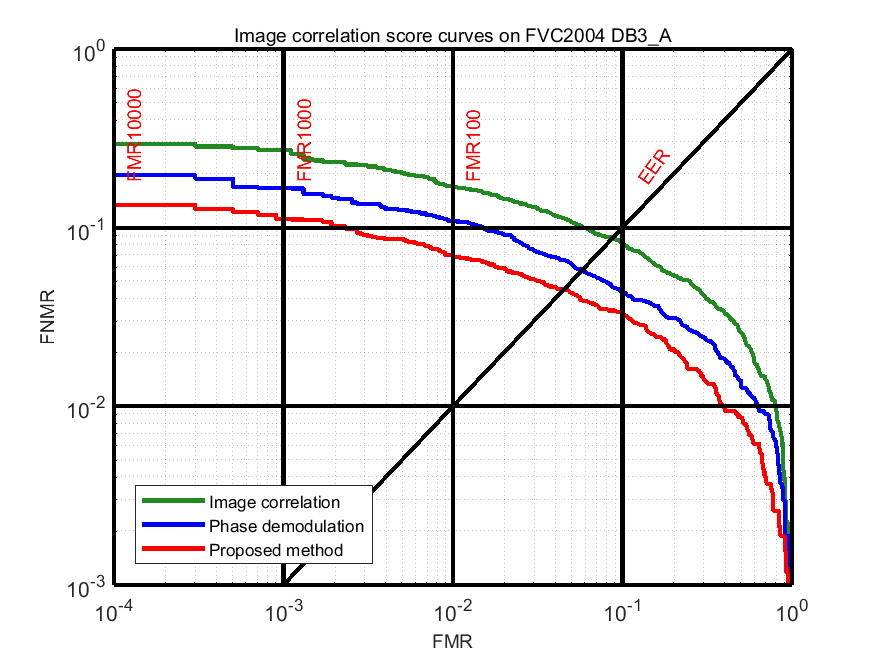}}
\centerline{(e) Image correlator on DB3{\_}A}\medskip
\end{minipage}
\hfill
\begin{minipage}[b]{.49\linewidth}
\centering
\centerline{\includegraphics[width=\linewidth]{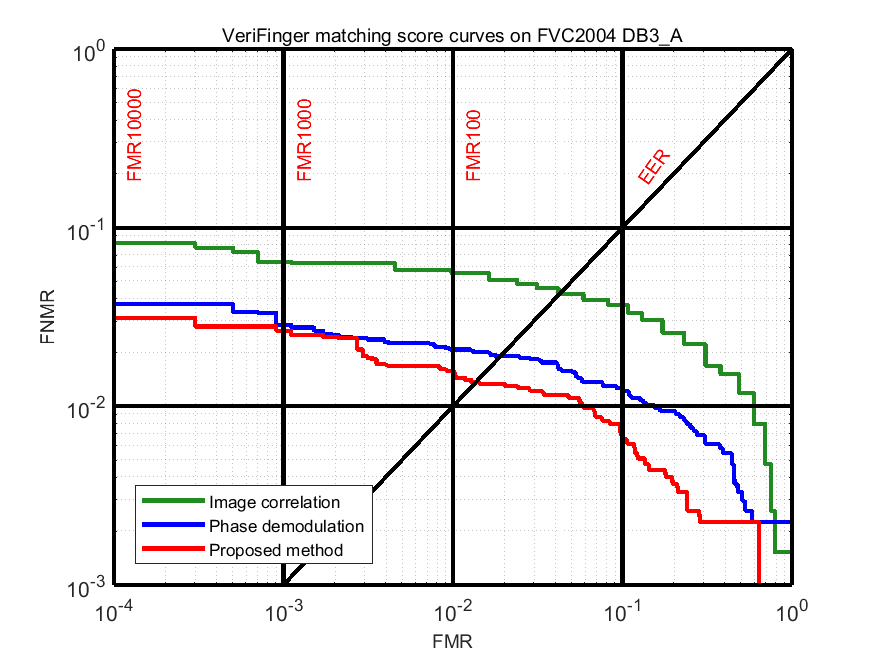}}
\centerline{(f) VeriFinger matcher on DB3{\_}A}\medskip
\end{minipage}
\caption{DET curves of image correlator and VeriFinger matcher with dense registration algorithms on FVC2004 DB1/2/3{\_}A.}
\label{fig:FVC_result}
\end{figure*}

\begin{figure}[!]
\begin{minipage}[b]{\linewidth}
  \centering
  \centerline{\includegraphics[width=\linewidth]{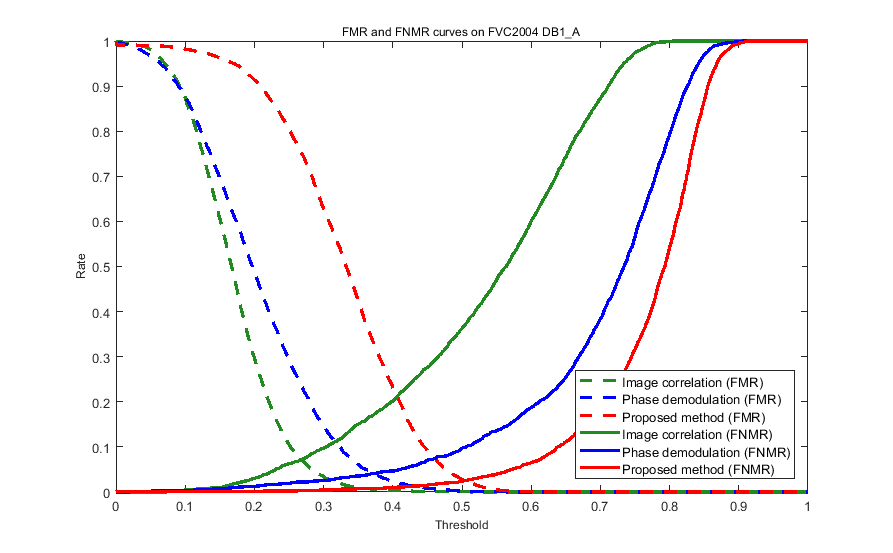}}
\end{minipage}
\caption{FMR and FNMR curves versus threshold of image correlator scores on FVC2004 DB1{\_}A using image correlation (green), phase demodulation (blue), and the proposed method (red).}
\label{fig:fmr}
\end{figure}

\begin{figure*}[!]
\begin{minipage}[b]{.49\linewidth}
  \centering
  \centerline{\includegraphics[width=\linewidth]{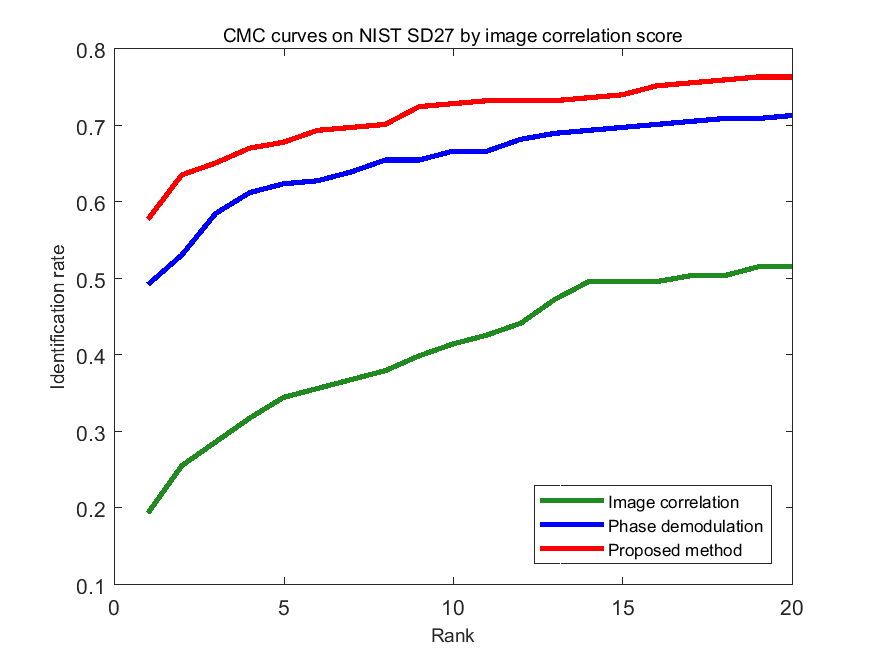}}
  \centerline{(a) Image correlator on NIST SD27}\medskip
\end{minipage}
\hfill
\begin{minipage}[b]{.49\linewidth}
  \centering
  \centerline{\includegraphics[width=\linewidth]{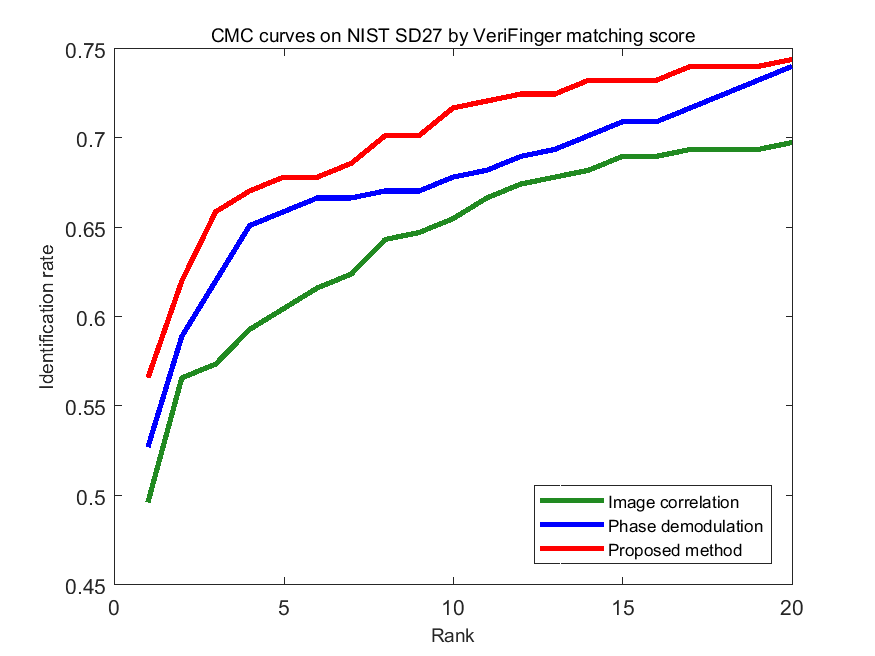}}
  \centerline{(b) VeriFinger matcher on NIST SD27}\medskip
\end{minipage}
\caption{CMC curves of image correlator and VeriFinger matcher with dense registration algorithms on NIST SD27.}
\label{fig:NIST_result}
\end{figure*}

\section{Experiment}\label{sec:experiment}
\subsection{Datasets}\label{ssec:datasets}
We evaluate the proposed algorithm and related dense registration methods \cite{Si2017}\cite{Cui2018} on multiple databases from aspects of registration accuracy, matching accuracy, mosaicking accuracy, and computing efficiency. Table \ref{table:database} provides a description of these databases. These databases were captured using different acquisition techniques (optical, capacitive, inking, and latent fingerprint development) and contain flat, rolled, and latent fingerprints. Various challenges such as distortion, low quality, and small area, are present in these databases.

\begin{table*}[!]
\caption{Fingerprint databases used in experiments.}  
\centering
\begin{tabular}{|c|c|c|c|c|c|} 
\hline
&FVC2004 DB1{\_}A& FVC2004 DB2{\_}A& FVC2004 DB3{\_}A& NIST SD27 &\tabincell{c}{Tsinghua Distorted\\Fingerprint Database\\(TDF)}\\
\hline
Image & \begin{minipage}{0.1\textwidth}\centerline{\includegraphics[scale=0.2]{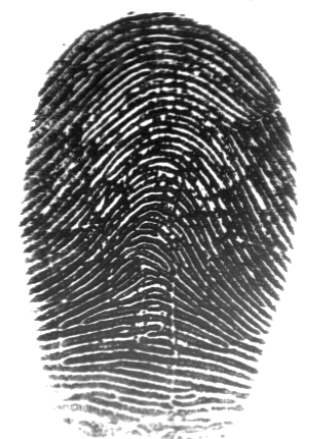}}\end{minipage} & \begin{minipage}{0.1\textwidth}\centerline{\includegraphics[scale=0.2]{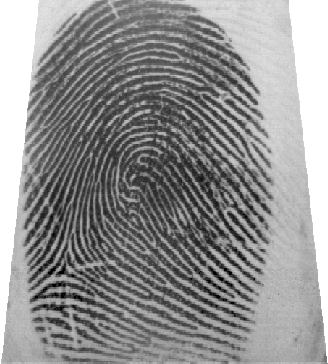}}\end{minipage} & \begin{minipage}{0.1\textwidth}\centerline{\includegraphics[scale=0.2]{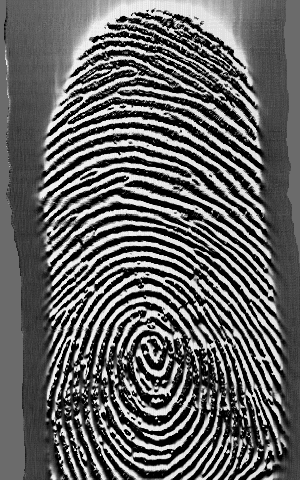}}\end{minipage} & \begin{minipage}{0.1\textwidth}\centerline{\includegraphics[scale=0.2]{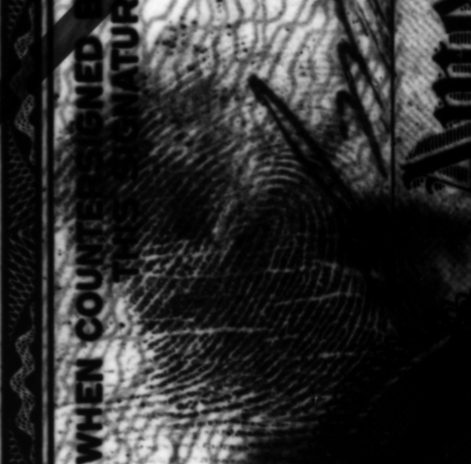}}\end{minipage} & \begin{minipage}{0.1\textwidth}\centerline{\includegraphics[scale=0.19]{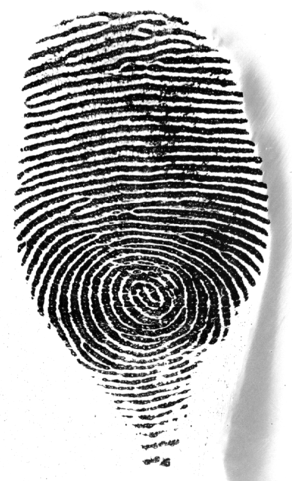}}\end{minipage}\\
\hline
Sensor & Optical & Optical & Thermal sweeping&Inking, latent &Optical\\
\hline
Description & \tabincell{c}{8 images$\times$100 fingers.\\Large distortion\\and various\\fingerprint quality}&\tabincell{c}{8 images$\times$100 fingers.\\Large distortion\\and various\\fingerprint quality} & \tabincell{c}{8 images$\times$100 fingers.\\Large distortion\\and various\\fingerprint quality} & \tabincell{c}{258 pairs of rolled\\and mated latent\\fingerprints of very\\low quality, small area,\\and distortion}&\tabincell{c}{320 pairs of highly\\distorted fingerprints,\\120 pairs with manually\\marked corresponding\\points}\\
\hline
Experiments &  \tabincell{c}{Registration accuracy\\Matching accuracy\\Mosaicking accuracy} & Matching accuracy & Matching accuracy &Matching accuracy& Registration accuracy\\
\hline
\end{tabular}
\label{table:database}
\end{table*}

\subsection{Registration Accuracy}\label{ssec:registration_accuracy}
We evaluate registration accuracy from two aspects: deviation of labeled corresponding points, and image correlation score. The former is a comparison between registration result with ground truth, while the latter can be tested on large dataset without labeled corresponding points. Finally, we show some good and poor registration examples.

\textbf{Evaluation by Deviation.} As a quantitative evaluation of registration accuracy, the registration errors are measured by calculating the deviation between labeled point correspondences. For example, for a pair of marked points $p_1$ in image 1 and $p_2$ in image 2, the algorithm translates $p_2$ in image 2 to a new position $\widetilde p$ after registration with image 1. Ideally, $\widetilde p$ should be equal to $p_1$. But in fact, there exists deviation between $\widetilde p$ and $p_1$, and $\left\| \widetilde p-p_1 \right\| _2$ is the registration error.

By calculating and analyzing the registration errors, we can get the cumulative distribution function (CDF) curves of registration errors in pixels on TDF database. Although TDF is involved in generating training data, it's used only for acquiring displacement fields, and the fingerprint images in TDF are not included during training. Additionally, the displacement field are automatically generated from distorted fingerprint video instead of manual labeling. Therefore, this evaluation is fair for different dense registration algorithms.

We use a subset of 120 pairs of fingerprints from TDF with manually marked corresponding points, which are also used for evaluating registration accuracy in previous dense registration methods \cite{Si2017}\cite{Cui2018}. As Fig. \ref{fig:reg_error} shows, the CDF curve of our method reaches 1 faster than previous methods. The cut-off error at rate 1 of our method is 13.5 pixels, which is smaller than image correlation (30.7) and phase demodulation (22.7). The mean registration errors of three dense registration methods are 5.65, 5.95, 3.75 pixels respectively, and our method reaches the minimum registration error.

\textbf{Evaluation by Image Correlator.} Successful registration reduces the geometric difference between different images from same fingerprint. In order to evaluate the registration accuracy on a larger scale, we conduct registration on the 2,800 pairs of genuine matching fingerprints from FVC2004 DB1{\_}A, and use the correlation coefficient to judge the registration accuracy. During the collection of three datasets of FVC2004, the impressions were required to vary in pressing pressure, skin distortion, and fingerprint quality \cite{Cappelli2006}. Therefore, we can further study the registration performances on different categories of fingerprints. Three types of fingerprints are selected by one of the authors manually, including 1) fingerprints with large distortion, 2) dry fingerprints, and 3) wet fingerprints. If a pair of genuine matching fingerprints involves one subset, its matching score is included in this subset. Some fingerprint pairs may involve two subsets, and their matching scores are counted into both subsets.

Fig. \ref{fig:fmr_2} shows the CDF curves of genuine matching scores using image correlator on different subsets on FVC2004 DB1{\_}A. The CDF curve of genuine matching score is also termed as False Non-Match Rate (FNMR) curve. At a fixed correlation score, a lower CDF value (cumulative probability) by image correlator indicates better registration result as well as lower FNMR. Comparing with phase demodulation method \cite{Cui2018}, the cumulative probability (or FNMR) at correlation score 0.5 by the proposed method is much lower: declined by 60.0\% on distorted fingerprints (subset 1), 81.3\% on dry fingerprints (subset 2), 76.0\% on wet fingerprints (subset 3), and 74.9\% on all fingerprints (whole set). Compared to image correlation method \cite{Si2017}, the proposed method produces even larger reduction in FNMR.

Fig. \ref{fig:fmr_2} shows that our method obviously outperforms previous dense registration methods within a wide range of correlation score ($\left[0.3, 0.8\right]$). The CDF curves of three dense registration methods only overlap at very low and very high correlation scores. But these two intervals are not important to compare the performances of three dense registration methods because: 1) A very low correlation score does not make sense because of the existence of very few failures on initial registration that all three dense registration methods will fail; 2) A very high correlation score also makes no sense because registration algorithm cannot eliminate the gray scale difference caused by image noise. That is to say, even if a pair of fingerprints are perfectly registered, their correlation coefficient cannot reach 1. Additionally, Fig. \ref{fig:fmr_2} shows that the proposed method successfully improves registration performances on distorted fingerprints (60.0\%) and low quality fingerprints (81.3\% and 76.0\%), which validates the purpose of this study to improve registration performance on fingerprints with distortion and low quality.

\textbf{Registration Success and Failure.} Some registration examples on FVC2004 DB1{\_}A by dense registration methods are displayed in Fig. \ref{fig:reg_examples}. Each row indicates a pair of input and reference fingerprints, as well as their registration results. Three dense registration methods are compared: 1) image correlation, 2) phase demodulation method, and 3) our method. Registration examples are displayed by overlapping reference fingerprint and aligned input fingerprint in colors to be easily understood. Green lines mark the well-registered areas, and red lines refer to registration failure regions or non-overlapping regions. Numbers inside the brackets are matching scores by VeriFinger matcher \cite{verifinger} and image correlator. Higher scores mean better registration performances, and the scores by image correlator directly reflect the accuracy of dense registration.

Those genuine matching examples in Fig. \ref{fig:reg_examples} show that the proposed method improves matching results with registration challenges. The first row is a pair of distorted fingerprints, the second row is an example of poor fingerprint quality, and the third row meets difficulties in aligning neighboring ridges in regions outside the fingerprint center because of lacking minutiae and self-similarity of ridge structures. But our method deals with those challenges well.

Meanwhile, as Fig. \ref{fig:failed_example} shows, our method still generates some bad registration results. These results mainly happen in situations where multiple challenges coexist, like small overlapping area, distortion, and low quality. Under these circumstances, the extracted minutiae are of small number or bad quality, and the minutiae-based initial registration result is far away from correct spatial transformation, which is beyond the range that dense registration method can handle. Previous dense registration methods \cite{Si2017}\cite{Cui2018} are also not able to solve this problem as they utilize the same minutiae-based initial registration as our method.

\subsection{Matching Accuracy}\label{ssec:matching_accuracy}
Matching experiments are conducted on FVC2004 DB1{\_}A, DB2{\_}A, DB3{\_}A, and NIST SD27 latent fingerprint database to show how registration process improves matching accuracy. Same as previous dense registration studies \cite{Si2017}\cite{Cui2018}, we use VeriFinger and image correlator to compute matching scores. Although there are many fingerprint matching methods \cite{2009handbook}, we select VeriFinger and image correlator because 1) they stand for minutiae-based matcher and image-based matcher respectively, 2) we are interested in relative increase in matching accuracy caused by registration algorithm, rather than the absolute performance, 3) they are easy to implement or access, and 4) using same matchers allows fair comparison with previous dense registration methods.

Fig. \ref{fig:FVC_result} shows the Detection Error Tradeoff (DET) curves on three FVC2004 subsets of dense registration methods: image correlation \cite{Si2017}, phase demodulation \cite{Cui2018}, and the proposed method. 
The subfigures (a)(c)(e) in Fig. \ref{fig:FVC_result} display the DET curves by image correlator, and (b)(d)(f) display the DET curves by VeriFinger matcher. It can be indicated from these plots that our method exceeds other dense registration methods on matching performances. 

To further study the matching performance in detail, we draw the False Match Rate (FMR) and False Non-Match Rate (FNMR) curves of image correlator scores of genuine and impostor matching by dense registration methods as Fig. \ref{fig:fmr}. As we can see, the proposed method increases correlation scores of genuine matching, which indicates a better registration performance. Additionally, the curves of impostor matching by our method has the smallest intersection region with genuine matching curves among all dense registration methods, which is beneficial for matching performance.

To further evaluate the proposed method on very low quality fingerprints, we performed matching experiments on NIST SD27. NIST SD27 contains 258 pairs of latent and mated rolled fingerprints, generating 258$\times$258 pairs of possible matching. The latent fingerprints are first enhanced by FingerNet \cite{fingernet2017}, then they are registered to rolled fingerprints by different dense registration algorithms. We compare our method with image correlation \cite{Si2017} and phase demodulation \cite{Cui2018}. The registered results are evaluated by VeriFinger and image correlator matcher. Fig. \ref{fig:NIST_result} shows the results of Cumulative Match Characteristic (CMC) curves. Our method outperforms other methods according to both matchers.

\begin{figure}[!]
\begin{minipage}[b]{.24\linewidth}
  \centering
  \centerline{\includegraphics[width=\linewidth]{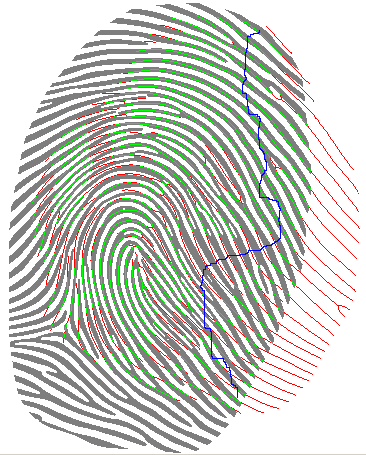}}
  \centerline{(a) Registration}
  \centerline{   Result}
\end{minipage}
\hfill
\begin{minipage}[b]{.24\linewidth}
  \centering
  \centerline{\includegraphics[width=\linewidth]{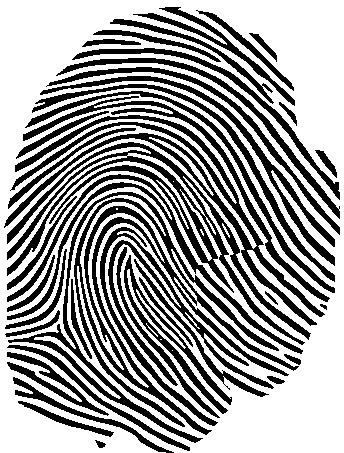}}
  \centerline{(b) Mosaicking}
  \centerline{   Result}
\end{minipage}
\hfill
\begin{minipage}[b]{.24\linewidth}
  \centering
  \centerline{\includegraphics[width=\linewidth]{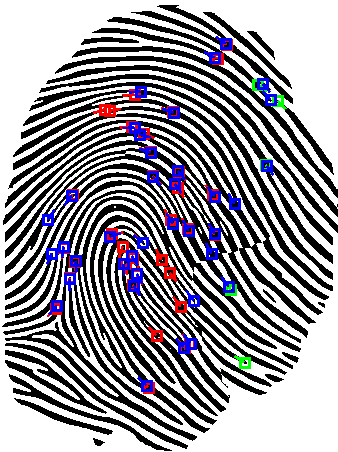}}
  \centerline{(c) Minutiae}
  \centerline{}
\end{minipage}
\caption{Fake minutiae from mosaicking error. The minutiae in red color are extracted from left part fingerprint before mosaicking, minutiae in green color are extracted from right part fingerprint before mosaicking, and minutiae in blue color are extracted after mosaicking. Several blue colored minutiae appear on mosaicking seam none-overlapping with red or green minutiae, and they are classified as `fake' minutiae.}
  \label{fig:mosaicking_minu}
\end{figure}
\subsection{Mosaicking Accuracy}\label{ssec:mosaic_accuracy}
A convenient way of quantifying the performances of fingerprint mosaicking is to evaluate minutiae extraction quality. The quality level of mosaicked fingerprint can be judged by counting `fake' and missing minutiae, i.e. the wrong minutiae caused by discontinuity from mosaicking. If the registration result is poor and misaligns some minutiae or ridges, the mosaicking result would have apparent discontinuity near the seam. Therefore, some fake minutiae appear at the location of ridge disconnection or displacement. 

Fig. \ref{fig:mosaicking_minu} (c) shows an example with fake minutiae in the overlapping area of two fingerprints. Minutiae in red color refer to the minutiae extracted from the left part fingerprint, and minutiae in green color refer to the minutiae from the right part fingerprint. Minutiae in blue color are extracted from mosaicked fingerprint, and some fake minutiae exist at seam location due to discontinuity. A fake minutia is defined as existing only on mosaicked fingerprint and neither of two fingerprints.

To examine the performance of the proposed fingerprint mosaicking as well as registration, we conduct minutiae extraction accuracy test on 2,800 genuine matching pairs from FVC2004 DB1{\_}A, and counts minutiae extraction errors resulted from mosaicking discontinuity on each pair. For a pair of genuine matching fingerprints, let $I_1$, $I_2$ and $I_M$ denote the input fingerprint, reference fingerprint, and mosaicked fingerprint, $\mathcal{R}_1$ and $\mathcal{R}_2$ are the overlapping region divided by the mosaicking seam. The minutiae extraction error is defined as:

\begin{equation}
e=\left| \widetilde{n}_1-n_1\right|+\left| \widetilde{n}_2-n_2\right|,
\label{equ:minutiae_error}
\end{equation}
where 
\begin{itemize}
\item $n_1=$ minutiae number in $I_1 (\mathcal{R}_1)$;
\item $n_2=$ minutiae number in $I_2 (\mathcal{R}_2)$;
\item $\widetilde{n}_1=$ minutiae number in $I_M (\mathcal{R}_1)$;
\item $\widetilde{n}_2=$ minutiae number in $I_M (\mathcal{R}_2)$.
\end{itemize}

\begin{figure}[htb]
\begin{minipage}[b]{1\linewidth}
  \centering
  \centerline{\includegraphics[width=.95\linewidth]{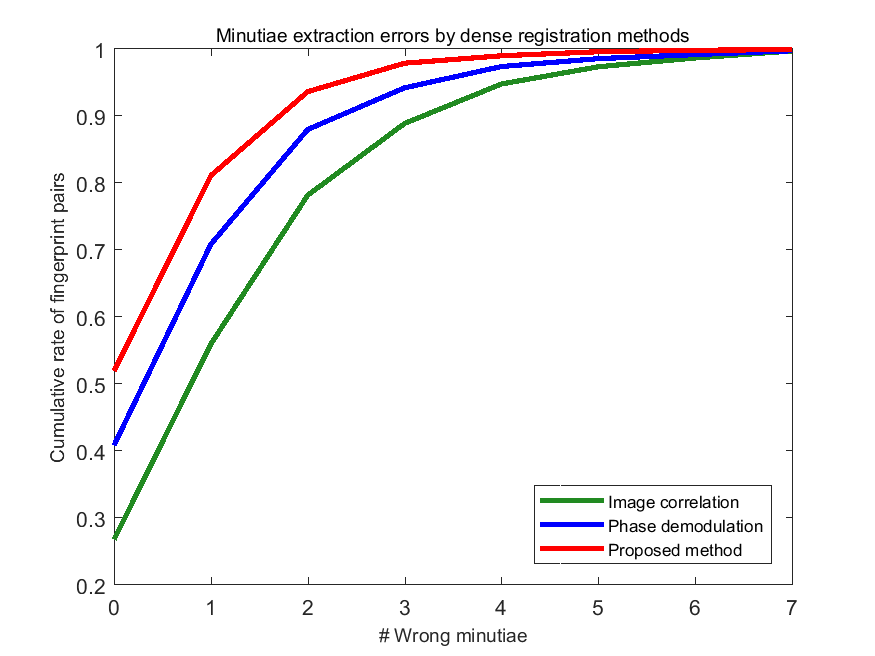}}
\end{minipage}
\caption{Minutiae extraction errors on 2,800 genuine matching pairs on FVC2004 DB1{\_}A by dense registration algorithms.}
\label{fig:mosaic_error}
\end{figure}

Because the fingerprints are already registered, it is reasonable to assume that all `real' minutiae of two fingerprints are aligned and matched. Therefore, directly counting minutiae numbers is enough to judge minutiae errors. In addition, we only conduct minutiae extraction on the region close to the mosaicking seam for efficiency. Because minutiae extraction error is resulted from bad mosaicking result, it only happens near mosaicking seam. 

The purpose of evaluating fingerprint mosaicking performances in this study is to judge the performances of dense registration algorithms. Therefore, we apply the proposed fingerprint mosaicking method after dense registration algorithms. We use the registration results of different dense registration methods to conduct fingerprint mosaicking. Fig. \ref{fig:mosaic_error} shows the cumulative curves of minutiae errors on FVC2004 DB1{\_}A caused by fingerprint mosaicking by image correlation \cite{Si2017}, phase demodulation \cite{Cui2018}, and the proposed method. Clearly, mosaicking results for our dense registration algorithm are better than previous methods, as more mosaicked fingerprints have no minutiae errors.

\subsection{Efficiency}\label{ssec:efficiency}
Table \ref{table:time} shows the average time cost to register a single pair of fingerprints by different dense registration algorithms on FVC2004 DB1{\_}A. The first two methods, image correlation \cite{Si2017} and phase demodulation \cite{Cui2018}, are implemented in C and MATLAB code, and tested on a CPU server equipped with Intel Xeon E5-2640 2.5GHz CPU. 

The average time cost of our method is 0.53s, including 0.38s for initial registration on CPU server which is the same as in \cite{Si2017} and \cite{Cui2018}, 0.15s for the dense registration on a GPU server with two Nvidia 1080Ti, and 2ms for nearest neighbor interpolation also on GPU, which is almost negligible. The dense registration part of our method is much faster than the counterpart of the other two methods. Meanwhile, different from previous dense registration methods that sample displacement field and fit a TPS transformation in the final step, our method directly interpolates to get a transformed image according to displacement field, which is more efficient.

\begin{table}[htb]
\caption{Average time costs (in seconds) of different dense registration algorithms for processing a pair of fingerprints in FVC2004 DB1{\_}A.}  
\centering
\begin{tabular}{|c|c|c|c|} 
\hline
Methods & \tabincell{c}{Image\\Correlation}& \tabincell{c}{Phase\\Demodulation} & \tabincell{c}{The Proposed\\Method} \\
\hline 
Time/s & 3 & 1.99 & 0.53\\
\hline 
\end{tabular}
\label{table:time}
\end{table}

\section{Conclusion}\label{sec:conclusion}
In this paper, we propose an end-to-end network to register fingerprints. Input fingerprints are first aligned by minutiae matching, then they are sent into the network to get a pixel-wise dense displacement field. Therefore, the input fingerprints are finely registered by outputted displacement field. We run registration and matching experiments on several databases and prove our registration method outperforms state-of-the-art dense registration method.

Comparing with previous dense registration methods of fingerprints, our method has two main advantages:
\begin{itemize}
\item [1)] By collecting and building training data from distorted and latent fingerprints, our method reaches the best registration and matching performances on various types of fingerprints.
\item [2)] By utilizing deep learning and training an end-to-end network to directly output displacement field, our method costs much less computation time than previous dense registration methods.
\end{itemize}

We also develop a fingerprint mosaicking method after registration by computing an optimal seam to stitch two fingerprints. Experiments on minutiae accuracy of mosaicked fingerprints testify our mosaicking method's performance, which is also a strong support of our registration method's superiority over other dense registration methods from mosaicking accuracy aspect.

Meanwhile, the bad registration examples in Fig. \ref{fig:failed_example} suggest that our method still needs improvement on registration accuracy. The current algorithm is a fine registration process which makes relatively small adjustment on initial registration result. Therefore, it suffers from very poor initial registration results. 

Future work will explore a more powerful algorithm that can handle long-range displacements and is able to correct initial registration errors. The computation speed of the proposed method also needs further improvement to be used in large-scaled fingerprint identification system.


\bibliographystyle{IEEEtran}
\bibliography{ref}
\vfill

\end{document}